\documentclass[10pt,twocolumn,letterpaper]{article}

\usepackage[pagenumbers]{wacv} 
\definecolor{wacvblue}{rgb}{0.21,0.49,0.74}
\usepackage[pagebackref,breaklinks,colorlinks,allcolors=wacvblue]{hyperref}

\usepackage{textcomp}
\usepackage{colortbl}
\usepackage{array,multirow}
\usepackage{booktabs}
\usepackage{xcolor}
\usepackage{tikz}
\usepackage{pifont}
\usepackage{bbold}
\usepackage{bm}
\usepackage{threeparttable}
\usepackage{booktabs}
\usepackage{tikz}
\usepackage{graphicx}
\usepackage{caption}
\usepackage{pgfplots}
\pgfplotsset{compat=1.18}
\usepackage{tcolorbox} 
\usepackage{multicol}
\usepackage{algorithm}
\usepackage{algpseudocode}
\usepackage{float}
\newcommand{\cmark}{\textcolor{green}{\ding{51}}}

\definecolor{lightblue}{rgb}{0.1, 0.4, 0.6}
\definecolor{ForestGreen}{rgb}{0.13, 0.65, 0.23}

\def\wacvPaperID{1437} 
\def\confName{WACV}
\def\confYear{2027}

\title{VLRC: Vision-Language Reprojection Consistency as a scalable signal for better feed-forward 3D pretraining}

\author{
Marwane Hariat$^{1}$ \quad David Filliat$^{2}$ \quad Antoine Manzanera$^{1}$\thanks{Corresponding author.}\\
$^{1}$U2IS, ENSTA -- Institut Polytechnique de Paris \quad
$^{2}$Agence Ministérielle pour l'IA de Défense\\
{\tt\small \{marwane.hariat, antoine.manzanera\}@ensta.fr \quad david.filliat@polytechnique.edu}
}

\begin{document}
\twocolumn[{
\maketitle
\begin{center}

\begin{minipage}[t]{0.48\textwidth}
\vspace{0pt} 
\raggedright

\begin{tikzpicture}

\node[anchor=south west] (main) at (0,0)
    {\includegraphics[width=\linewidth]{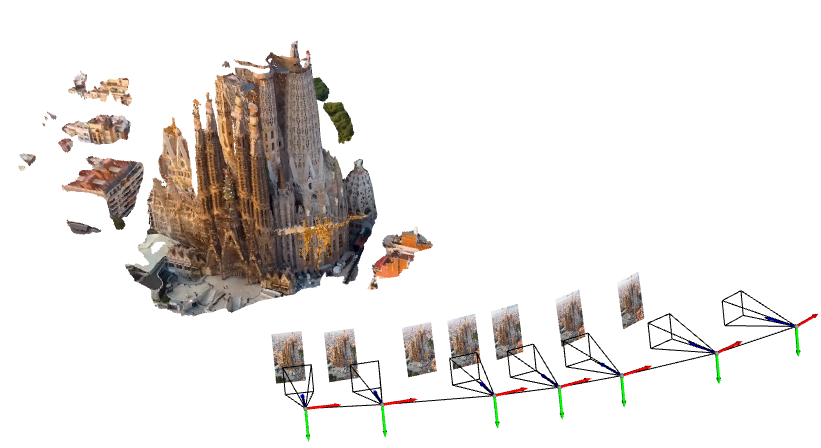}};

\node (img1) at (4.85, 3.2) {\includegraphics[width=0.075\textwidth, height=0.1\textwidth]{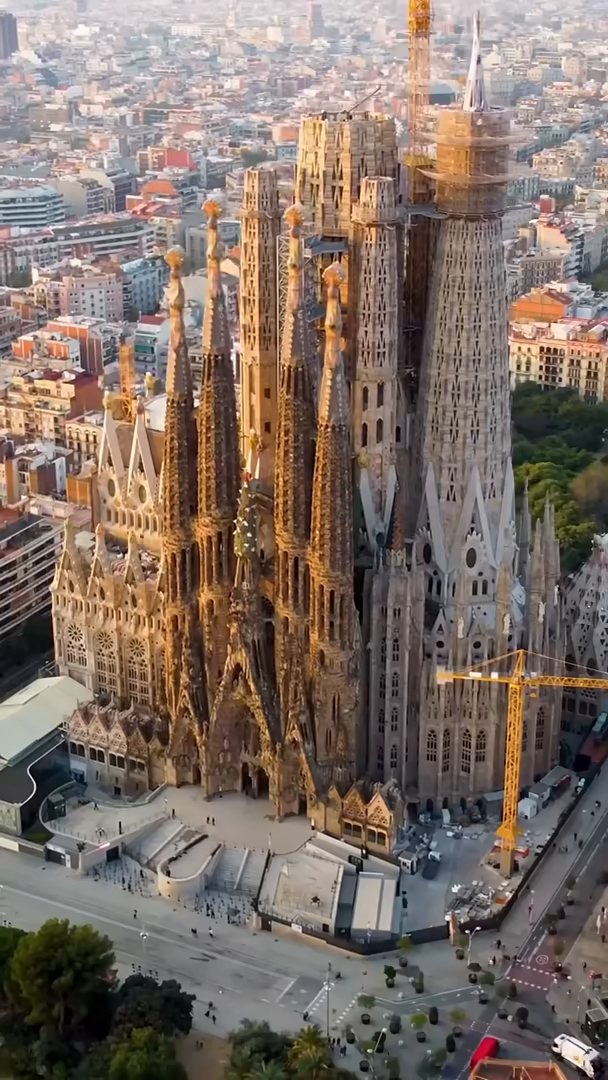}};
\node (img2) at (5.49, 3.2) {\includegraphics[width=0.075\textwidth, height=0.1\textwidth]{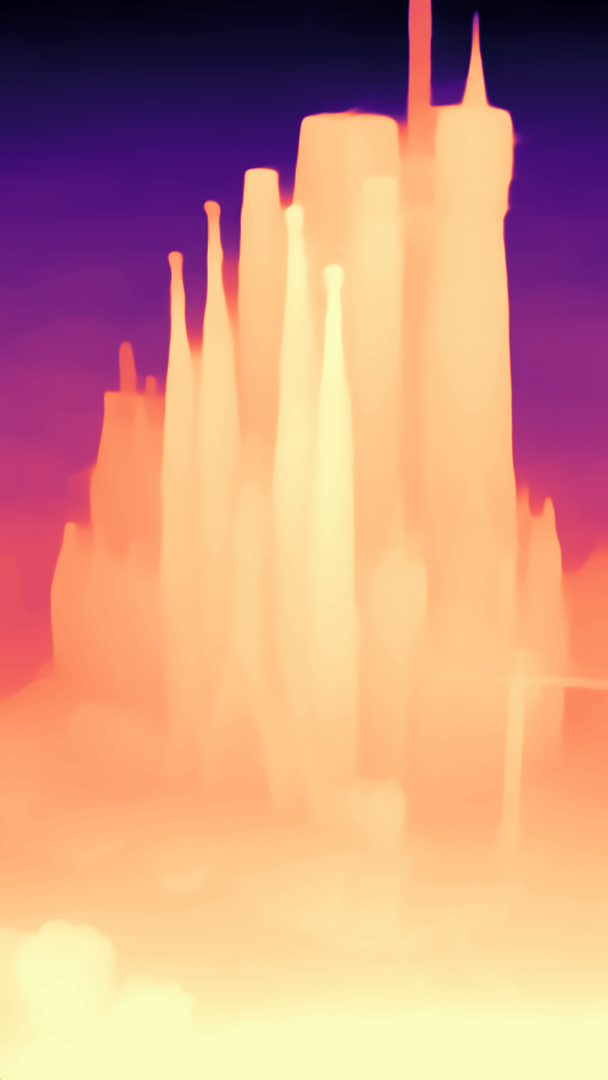}};

\node (img1) at (7.0, 3.2) {\includegraphics[width=0.075\textwidth, height=0.1\textwidth]{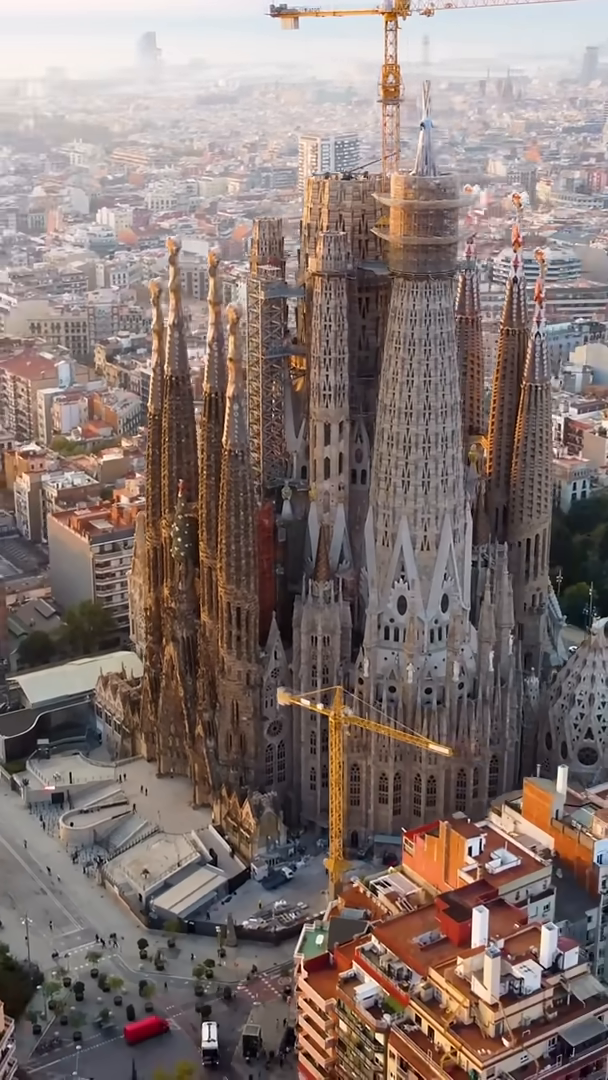}};
\node (img2) at (7.64, 3.2) {\includegraphics[width=0.075\textwidth, height=0.1\textwidth]{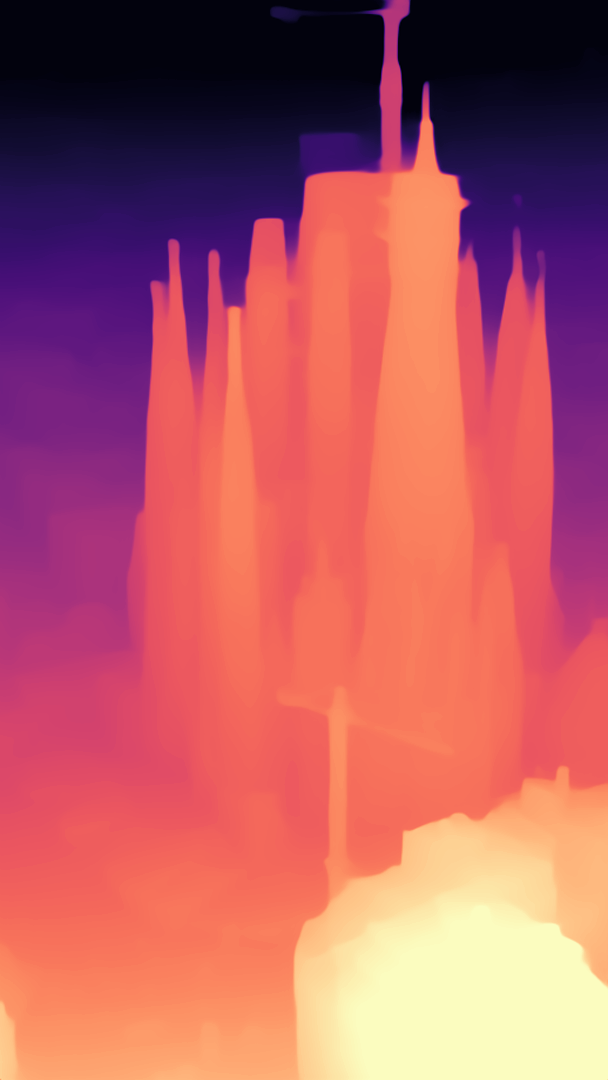}};

\coordinate (p1) at (2.9, 1.3);
\coordinate (target1) at (4.5, 2.8);
\draw[dotted, line width=1.5pt, color=violet!80!black] (p1) -- (target1);

\coordinate (p2) at (3.2, 1.3);
\coordinate (target2) at (5.1, 2.8);
\draw[dotted, line width=1.5pt, color=violet!80!black] (p2) -- (target2);

\coordinate (p3) at (6.45, 1.7);
\coordinate (target3) at (6.7, 2.8);
\draw[dotted, line width=1.5pt, color=cyan!70!green] (p3) -- (target3);

\coordinate (p4) at (6.7, 1.9);
\coordinate (target4) at (7.3, 2.8);
\draw[dotted, line width=1.5pt, color=cyan!70!green] (p4) -- (target4);

\end{tikzpicture}

\end{minipage}%
\hfill
\begin{minipage}[t]{0.48\textwidth}
\vspace{0pt} 
\raggedright

\begin{tikzpicture}

\node (img1) at (1.5,2.5) {\includegraphics[width=0.3\textwidth, height=0.3\textwidth]{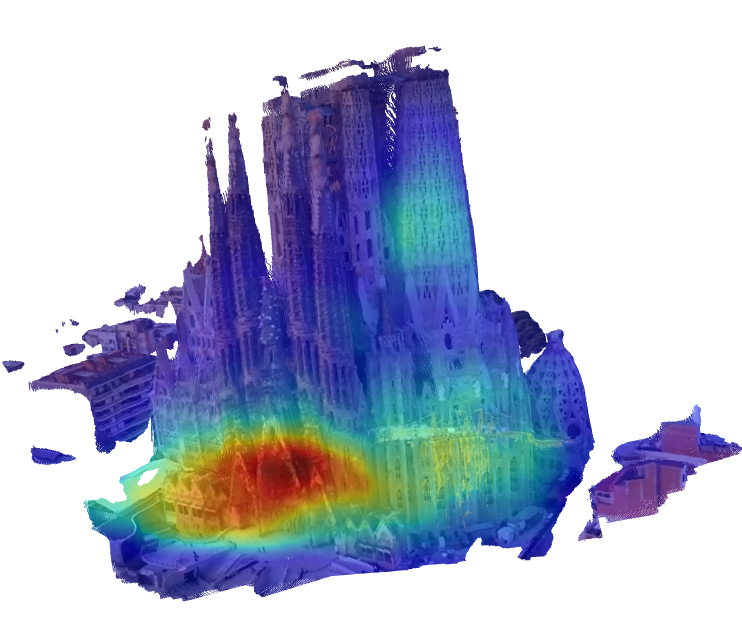}};
\node (img2) at (5.15, 1.45) {\includegraphics[width=0.3\textwidth, height=0.65\textwidth]{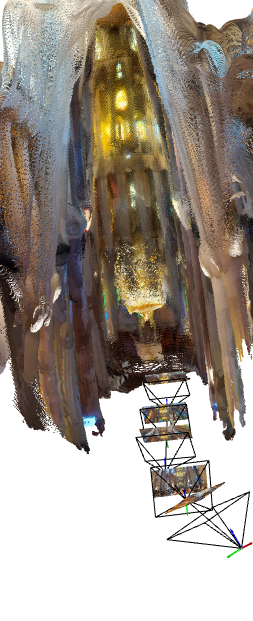}};
\node (img3) at (7.5, 3.2) {\includegraphics[width=0.15\textwidth, height=0.2\textwidth]{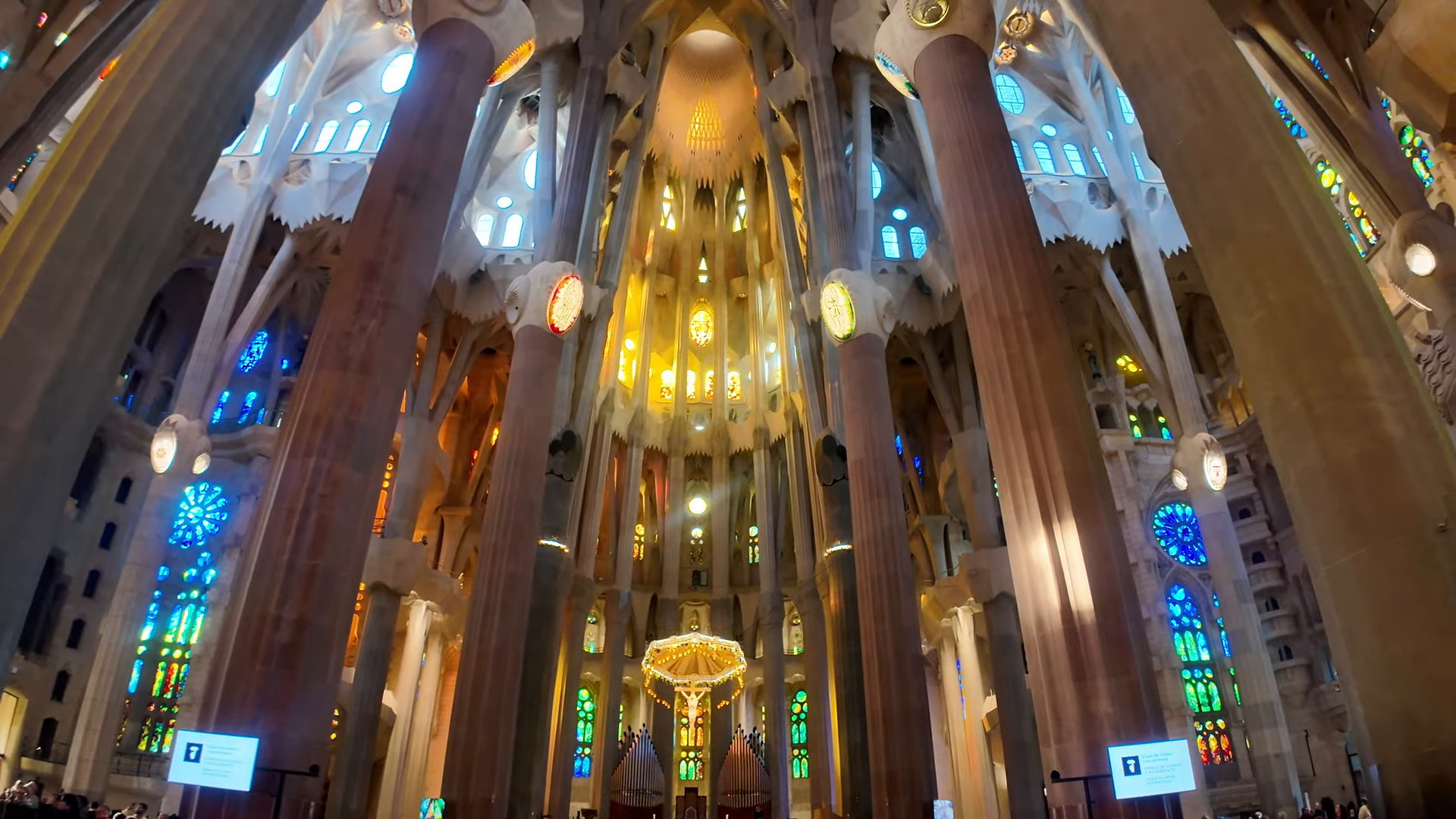}};
\node (img4) at (7.5, 1.2) {\includegraphics[width=0.15\textwidth, height=0.2\textwidth]{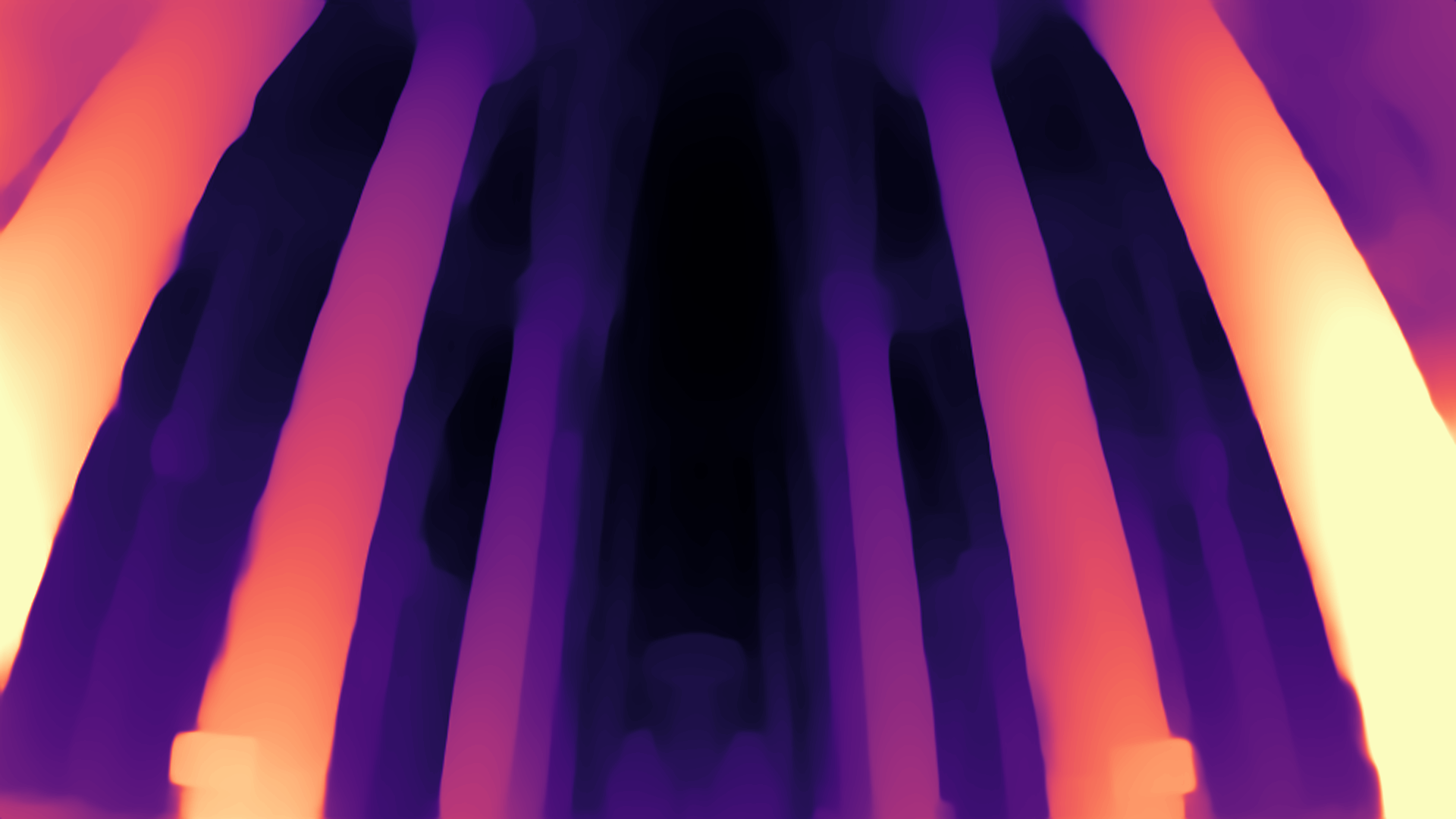}};

\node[black, font=\small\bfseries, align=center] at (1.5,0.75) {Show me the entrance \\ of the cathedral};



\coordinate (p1) at (5.7, 0.6);
\coordinate (target1) at (6.9, 4);
\draw[dotted, line width=1.5pt, magenta!80!black] (p1) -- (target1);

\coordinate (p2) at (5.8, 0.3);
\coordinate (target2) at (6.9, 2.4);
\draw[dotted, line width=1.5pt, magenta!80!black] (p2) -- (target2);

\end{tikzpicture}

\end{minipage}%
\vspace{-15pt}
\captionof{figure}{
\textbf{SS3D+VLRC trained on YouTube8M web videos}
reconstructs both the exterior (left) and interior (right) of the Sagrada Familia from two casual videos: one recorded outside and one inside. It accurately localizes the entrance in 3D using the prompt (“Show me the entrance of the cathedral”). The reconstruction uses self-supervised estimates of depth, camera pose, and intrinsics. Depth maps and camera trajectories are visualized, 
each camera 
being
shown along 
with
its corresponding viewpoint.
Point clouds are rendered with Open3D.}
\label{fig:sagrada}
\end{center}
}]

\begin{abstract}
Feed-forward 3D models are commonly trained using either expensive geometric supervision or self-supervised photometric objectives, both of which provide incomplete learning signals. We introduce Vision-Language Reprojection Consistency (VLRC), a scalable auxiliary objective that exploits frozen vision-language representations as semantic multi-view supervision. Given a predicted 3D reconstruction, VLRC reprojects dense vision-language features across views and enforces feature consistency between corresponding image locations, requiring no additional 3D annotations. The objective integrates seamlessly with both self-supervised monocular reconstruction and supervised-pretrained feed-forward 3D models during unlabeled adaptation. By aligning geometry with language-grounded features, VLRC not only improves depth and camera estimation but also enables more coherent multi-view semantic fusion for open-vocabulary 3D scene understanding. Experiments on indoor and outdoor benchmarks demonstrate consistent gains in 3D reconstruction accuracy and zero-shot open-vocabulary 3D semantic segmentation.

\end{abstract}    
\section{Introduction}
\label{sec:intro}

Estimating 3D structure from monocular videos is a central problem in computer vision. Recent large-scale feed-forward 3D models have shown that neural networks can directly predict rich geometric quantities, including depth, camera motion, intrinsics, and 3D structure. A first line of work relies on large-scale geometric supervision, such as depth, camera pose, intrinsics, or point clouds. Methods such as Depth Anything~\cite{lin2025depth}, VGGT~\cite{wang2025vggt}, \(\pi^3\)~\cite{wang2025pi}, DUSt3R~\cite{wang2024dust3r}, MapAnything~\cite{keetha2025mapanything}, and CUT3R~\cite{wang2025continuous} leverage large amounts of annotated or reconstructed 3D data, such as ground-truth depth, calibrated multi-view captures, LiDAR scans, synthetic data, or pseudo-labels distilled from stronger reconstruction systems. This form of supervision enables impressive feed-forward 3D prediction across diverse benchmarks, but it has several drawbacks: it remains expensive to obtain, can inherit biases from the datasets, sensors, reconstruction pipelines, or teacher models used to generate it, and is difficult to extend continually as models encounter new in-the-wild video domains.

A second line of work aims to reduce this dependence on explicit 3D supervision through self-supervised learning from monocular video. In particular, SS3D~\cite{hariat2026ss3d} shows that a single feed-forward model can jointly learn depth, camera motion, and intrinsics from large-scale raw web videos using structure-from-motion reprojection losses. Such self-supervised objectives are attractive because they can scale to unannotated videos, but they are appearance-driven and can become ambiguous under illumination changes, low-texture regions, repeated patterns, specularities, motion blur, occlusions, and dynamic objects.

Despite their differences, both supervised and self-supervised feed-forward 3D models rely on incomplete training signals: supervised training is limited by the cost, bias, and limited continual scalability of annotations or pseudo-labels, while self-supervised training is limited by the ambiguity of photometric consistency. This motivates an additional scalable signal that can complement both forms of 3D pretraining.

Vision-language models such as CLIP~\cite{radford2021learning,zhou2022extract,luddecke2022image}, SigLIP~\cite{zhai2023sigmoid,tschannen2025siglip}, and VL-JEPA~\cite{chen2025vl} provide a promising source of such a signal. By aligning images with language at web scale, they learn visual features enriched with language-grounded semantic priors beyond visual appearance. These features have been widely used for open-vocabulary 3D semantic segmentation, but they are typically lifted and fused only after the 3D structure has been estimated. As a result, the quality of the 3D semantic representation depends directly on the quality of the predicted 3D structure: if the 3D prediction is inaccurate, the same physical point may receive inconsistent vision-language features across views, leading to noisy 3D semantics.

Our key observation is that this inconsistency can itself be used as a training signal. If a predicted 3D structure induces inconsistent dense vision-language features across multiple views of the same scene, then the predicted geometry is likely misaligned with the underlying 3D scene structure. Enforcing this consistency provides a scalable signal that complements both supervised and self-supervised 3D pretraining: it does not require ground-truth 3D annotations, and it is less tied to low-level RGB appearance than photometric reprojection, providing useful gradients in ambiguous regions such as textureless surfaces or illumination-varying areas. In addition, we show that this signal can be integrated into existing unlabeled post-training pipelines, such as SelfEvo-style adaptation \cite{huang2026self}, where it provides an additional vision-language consistency term for adapting pretrained 3D models to new video domains.

We introduce \textbf{Vision-Language Reprojection Consistency} (VLRC), a general auxiliary objective for large-scale feed-forward 3D pretraining. Given a model that predicts 3D structure from monocular videos, VLRC uses the induced 3D reconstruction to reproject dense vision-language-aligned features across views and enforces multi-view consistency in feature space. This objective requires no additional 3D annotations. Instead, it reuses frozen vision-language representations as a scalable feature-space signal that complements existing 3D training objectives. VLRC improves 3D learning in two complementary ways. First, it adds a feature-space supervision signal for 3D estimation: the predicted 3D structure must explain not only RGB appearance or available 3D targets, but also the multi-view consistency of high-level vision-language features. Second, by enforcing this consistency, VLRC aligns the predicted 3D structure with dense semantic representations, enabling more coherent fusion of 2D vision-language features into 3D. As a result, VLRC improves both core 3D estimation and downstream 3D semantic understanding.

To summarize, our main contributions are:
\begin{enumerate}
\item We introduce \textbf{Vision-Language Reprojection Consistency} (VLRC), a general auxiliary objective that uses predicted 3D structure to enforce multi-view consistency of dense vision-language features.

\item We demonstrate that VLRC improves core 3D estimates, including depth, camera motion, intrinsics, and induced 3D reconstruction, across multiple datasets and across both self-supervised and supervised feed-forward 3D settings.

\item We show that VLRC produces 3D structure that is better aligned with dense vision-language features, improving downstream zero-shot open-vocabulary 3D semantic segmentation.

\end{enumerate}

\section{Related work}
\label{sec:related_Works}

\paragraph{Large-scale supervised feed-forward methods.}
Recent progress in feed-forward 3D estimation has been driven by large-scale supervision from depth, camera calibration, reconstructed point clouds, or teacher-generated pseudo-labels. MiDaS~\cite{ranftl2020towards} showed that mixing heterogeneous labeled depth datasets with scale- and shift-invariant losses yields robust zero-shot relative depth. Depth Anything~\cite{yang2024depth,lin2025depth} further scaled monocular depth learning by combining labeled data with pseudo-labels generated on large unlabeled image collections, using strong visual backbones such as DINOv2~\cite{oquab2023dinov2}. Beyond monocular depth, DUSt3R~\cite{wang2024dust3r} predicts dense point maps from image pairs using cross-view transformer reasoning initialized from CroCo~\cite{Weinzaepfel22CroCo}, while MASt3R~\cite{leroyMast3R2024} adds dense matching features for improved correspondence and reconstruction. More recent feed-forward models predict richer 3D structure from multiple views: VGGT~\cite{wang2025vggt} jointly predicts cameras, depth, point maps, and tracking features; MapAnything~\cite{keetha2025mapanything} accepts optional geometric inputs such as intrinsics, poses, or depth; and CUT3R~\cite{wang2025continuous} introduces a continuous reconstruction formulation with memory for temporally coherent video reconstruction.

These supervised or pseudo-supervised models demonstrate the power of large-scale 3D targets, but they remain dependent on annotated datasets, reconstruction pipelines, sensors, or teacher models. Their predictions may inherit the biases of these sources, and adapting them continually to new in-the-wild video domains can require additional target generation, as observed in SelfEvo~\cite{huang2026self}. Our work is complementary: rather than introducing a new feed-forward architecture, VLRC provides an auxiliary vision-language reprojection signal that can be added to existing 3D models to improve geometry and support unlabeled adaptation.

\paragraph{Self-supervised and continual feed-forward 3D adaptation.}
Self-supervised monocular 3D learning estimates depth, camera motion, and intrinsics by synthesizing one frame from another and comparing it to the observed image, following the view-synthesis formulation popularized by Zhou et al.~\cite{zhou2017unsupervised}. This removes the need for ground-truth 3D, but the supervision remains primarily photometric and is therefore ambiguous under illumination changes, low texture, repeated patterns, specularities, motion blur, occlusions, and dynamic objects. Many works address specific violations of this assumption: Marsal et al.~\cite{marsal2023brightflow} model brightness changes; Godard et al.~\cite{godard2019digging} use minimum reprojection over multiple source frames to handle occlusions; Li et al.~\cite{li2021unsupervised} estimate residual 3D motion for dynamic objects; Hariat et al.~\cite{hariat2023rebalancing} identify moving regions using discrepancies between optical flow and depth-based reprojection; and Shu et al.~\cite{shu2020feature} and Hariat et al.~\cite{hariat2025improved} introduce feature- or contour-based cues to better supervise weakly textured regions.

Although effective, these mechanisms are often specialized to individual failure modes. VLRC instead provides a unified feature-space signal: when predicted depth, pose, or intrinsics induce inconsistent dense vision-language features across views, the model receives a reprojection penalty in a representation space that is less tied to raw RGB appearance. This is motivated by recent evidence that vision-language-aligned features are sensitive to multi-view geometric inconsistencies~\cite{asim2025met3r}. Since these features are learned from large-scale image-text data, they encode higher-level semantic cues that complement photometric consistency.

A related line of work studies continual learning~\cite{fini2022self} and self-distillation~\cite{caron2021emerging,he2022masked} for annotation-free adaptation. In feed-forward 3D reconstruction, SelfEvo~\cite{huang2026self} adapts a pretrained multi-view model such as VGGT on unlabeled videos: a teacher processes the full input sequence and produces stop-gradient reconstruction targets, while a student receives a reduced-context sequence and learns to recover the same reconstruction; the teacher is updated as an exponential moving average of the student. VLRC is complementary to this strategy, providing an additional vision-language reprojection signal that can be combined with SelfEvo-style post-training.

\paragraph{Open-vocabulary 3D semantic segmentation.}
3D scene understanding has moved from closed-set geometric recognition toward open-vocabulary representations built from large-scale vision-language models. Early 3D recognition methods such as VoteNet~\cite{qi2019deep} reason directly on point clouds using deep point-set features and Hough voting. Recent open-vocabulary methods instead transfer 2D vision-language features into 3D. PLA~\cite{ding2023pla} and PartSLIP~\cite{liu2023partslip} project 3D points into multiple views, extract CLIP features, and aggregate them into point-level representations. PointCLIP~\cite{zhang2022pointclip} learns view aggregation and selection strategies, related to multi-view selection schemes such as~\cite{song2025mv}. OpenScene~\cite{peng2023openscene} fuses multi-view image features and distills them into a 3D network, while later works use image- or region-level descriptions from large vision-language models to associate language with 3D points~\cite{jiang2024open,zhou2025ov3d}. Casper3D~\cite{hariat2026lightweight} further converts noisy multi-view 2D foundation-model embeddings into latent 3D semantic representations using a Bayesian inverse strategy.

Most of these methods assume reliable geometry, such as ground-truth depth, calibrated poses, intrinsics, or high-quality reconstructed point clouds. This assumption is not guaranteed when geometry is predicted by feed-forward 3D models from monocular videos: a reconstruction may be sufficient for RGB reprojection while still producing inconsistent multi-view vision-language features. VLRC addresses this issue during 3D training by encouraging the predicted geometry to align dense vision-language features across views, enabling more coherent 2D-to-3D feature lifting and stronger text-queryable 3D semantic segmentation.

\section{Method}
\label{sec:method}

\subsection{Problem setup}
\label{sec:method:problem_setup}

Let \(f_{\theta}\) be a pretrained feed-forward multi-view reconstruction model
\[
f_\theta : \{I_i\}_{i=1}^{N} \longrightarrow \{(D_i, T_i, K_i)\}_{i=1}^{N}
\]
that maps a short clip of $N$ consecutive RGB frames to per-frame depth, pose, and intrinsics. We denote:

\begin{itemize}
    \item $D_t(p) \in \mathbb{R}_{+}$: the predicted depth at pixel $p$,
    \item $T_t =
\begin{bmatrix}
R_{t} & \mathbf{t}_{t} \\
\mathbf{0} & 1
\end{bmatrix}$: the camera pose to world with $R_t \in SO(3)$ and \(\mathbf{t}_t \in \mathbb{R}^3\),
    \item $K_t \in \mathbb{R}^{3\times 3}$: the camera intrinsics matrix.
\end{itemize}

\paragraph{3D structure.}
These predictions induce a 3D reconstruction of the scene. Let \(p=(u,v,1)^\top\) be a homogeneous pixel coordinate. A pixel \(p\) is first back-projected to 3D in camera coordinates as
\begin{equation}
X_t(p) = D_t(p) K_t^{-1} p\quad \in \mathbb{R}^3.
\label{eq:backproject_3d}
\end{equation}

It is then transformed into world coordinates as:
\begin{equation}
Y_t(p)=R_tX_t(p)+\mathbf{t}_t.
\end{equation}

Aggregating such points over frames yields the predicted 3D reconstruction.

\paragraph{Supervised 3D pre-training.} In a supervised feed-forward 3D setting, the model is trained using ground-truth or pseudo-ground-truth 3D targets. Let \(D_i^{\star}\), \(T_i^{\star}\), and \(K_i^{\star}\) denote the target depth, camera pose, and intrinsics for frame \(i\). For notational simplicity, we write the multi-task loss terms using a generic robust discrepancy \(\|\cdot\|_{\epsilon}\). In practice, the exact form of each term is model-dependent and may include scale-invariant, shift-invariant, normalized, or task-specific variants.

\begin{equation}
\begin{aligned}
\mathcal{L}_{\mathrm{3D}}^{\mathrm{sup}}
&=
\sum_{i=1}^{N}
\left\|
\mathbf{D}_i - \mathbf{D}_i^{\star}
\right\|_{\epsilon}
+
\sum_{i=1}^{N}
\left\|
\mathbf{T}_i - \mathbf{T}_i^{\star}
\right\|_{\epsilon}
+
\sum_{i=1}^{N}
\left\|
\mathbf{K}_i - \mathbf{K}_i^{\star}
\right\|_{\epsilon}
\\
&=
\mathcal{L}_{\mathrm{depth}}
+
\mathcal{L}_{\mathrm{camera}}
+
\mathcal{L}_{\mathrm{intr}} .
\end{aligned}
\label{eq:3d_loss_supervision}
\end{equation}

Additional supervised terms, such as point-map or tracking losses, can be included when they are part of the base reconstruction model. 

\paragraph{Self-supervised training.} In the self-supervised setting, the model uses predicted depth, camera motion, and intrinsics to warp a nearby source frame \(I_s\) into the coordinate frame of a target frame \(I_t\). The pixel \(p_t\) is first back-projected in the target camera coordinate system as in Eq~\ref{eq:backproject_3d}, transformed to the source camera, and then projected into the source image. The resulting sampled source color gives the synthesized target image \(\hat I_t\), which is compared to the observed target image \(I_t\):

\begin{equation}
\begin{aligned}
T_{t\rightarrow s} &=
\begin{bmatrix}
R_{t\rightarrow s} & \mathbf{t}_{t\rightarrow s} \\
\mathbf{0} & 1
\end{bmatrix}
\in SE(3), \\[1mm]
X_t(p_t) &= D_t(p_t)K_{t}^{-1}p_t, \\[1mm]
p_s &=
\pi\!\left(
K_{s} T_{t\rightarrow s}
\begin{bmatrix}
X_t(p_t) \\
1
\end{bmatrix}
\right), \\[1mm]
\hat I_{t}(p_t)
&=
I_s(p_s), \\[1mm]
\mathcal{L}_{\mathrm{photo}}
&=
\sum_{p_t\in\Omega}
\rho\!\left(
\hat I_{t}(p_t)-I_t(p_t)
\right),\\[1mm]
\mathcal{L}_{\mathrm{3D}}^{\mathrm{self}}
&= \mathcal{L}_{\mathrm{photo}} + \mathcal{L}_{\mathrm{reg}}
\end{aligned}
\label{eq:3d_loss_self}
\end{equation}

where \(\mathcal{L}_{\mathrm{reg}}\) denotes regularization terms such as depth smoothness or other model-specific priors.

Here, \(T_{t\rightarrow s} = T_{s}^{-1}T_{t}\in SE(3)\) is the relative transformation from the target camera to the source camera, \(\pi(\cdot)\) denotes perspective projection, and \(I_s(p_s)\) is obtained with differentiable bilinear sampling. The function \(\rho(\cdot)\) is a robust photometric penalty, typically combining an RGB 
comparison
term with an SSIM comparison term. This objective encourages the predicted depth, pose, and intrinsics to explain the target image through view synthesis in color space.

We use \(\mathcal{L}_{\mathrm{3D}}\) to denote the base reconstruction objective, either \(\mathcal{L}_{\mathrm{3D}}^{\mathrm{sup}}\) defined in Eq.~\ref{eq:3d_loss_supervision}, \(\mathcal{L}_{\mathrm{3D}}^{\mathrm{self}}\) defined in Eq.~\ref{eq:3d_loss_self}, or another model-specific objective. In Sec.~\ref{sec:experiments}, we instantiate this objective in two settings: self-supervised video pretraining with SS3D~\cite{hariat2026ss3d} and supervised feed-forward reconstruction with VGGT~\cite{wang2025vggt}.

\begin{figure*}[ht]
  \centerline{\includegraphics[width=0.9\linewidth]{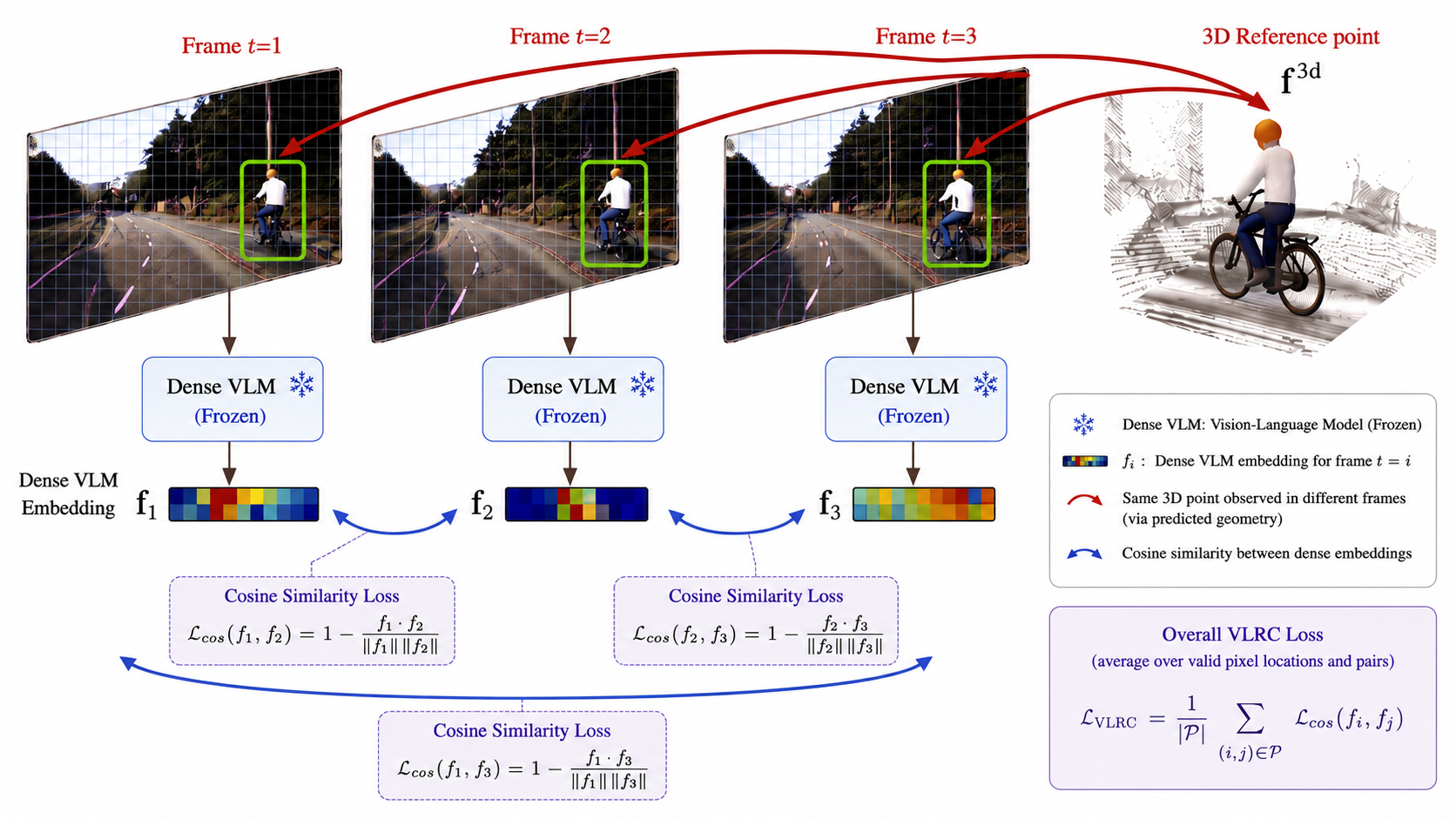}}
  \caption{{\bf Reprojection of a point \(p \in \mathbb{R}^3\): the head of a cyclist} - The 3d point centered on the head of the cyclist is reprojected on the successive frames and the corresponding Dense VLM features are extracted with a frozen encoder and compared across the reprojected correspondences using a cosine dissimilarity loss. VLRC encourages the predicted geometry to align multi-view language-grounded features consistently across views. \(\mathcal{P}\) denotes the set of target-source frame pairs used for reprojection.}
  \label{fig:reprojection_vlrc}
\end{figure*}

\subsection{Vision-Language Reprojection Consistency}
\label{sec:method:clip_reprojection}
The reprojection formulation in Eq.~\ref{eq:3d_loss_self} enforces RGB consistency through photometric losses. More generally, the same geometry-induced correspondences can be applied to dense vision-language-aligned features. We call this auxiliary signal \textbf{Vision-Language Reprojection Consistency}.

Let \(F_i=\phi_{\mathrm{VLM}}(I_i)\) be a dense vision-language aligned feature map extracted from frame \(I_i\). Using the same projected source coordinate \(p_s\)
as in Eq.~\ref{eq:3d_loss_self}, 
we synthesize the target feature map as:
\[
\hat F_{t}(p_t)=F_s(p_s).
\]

We then enforce feature-level consistency between \(\hat F_{t}\) and \(F_t\), as explained on Fig.~\ref{fig:reprojection_vlrc}, yielding the final objective:

\begin{equation}
\mathcal{L}_{\mathrm{VLRC}}
=
\sum_{(t,s)\in\mathcal{P}}
\sum_{p_t\in\Omega}
m_{t,s}(p_t)
\left[
1-
\cos\!\left(
\hat F_{t}(p_t),
F_t(p_t)
\right)
\right],
\label{eq:vlrc_reprojection_loss}
\end{equation}

\begin{equation}
\mathcal{L}_{\mathrm{final}}
=
\mathcal{L}_{\mathrm{3D}}
+
\lambda_{\mathrm{VLRC}}\mathcal{L}_{\mathrm{VLRC}},
\label{eq:final_loss}
\end{equation}
where \(\Omega\) is the set of pixels, \(\lambda_{\mathrm{VLRC}}\) controls the strength of the vision-language reprojection signal. Gradients are back-propagated through the differentiable reprojection operation into the 3D model, while the vision-language encoder remains frozen.
Here \(m_{t,s}\) masks invalid projections and unreliable correspondences. Additional details are provided in Sec.~\ref{sec:experiments:implementation_details}.

\subsection{Open-Vocabulary 3D Semantic Segmentation}
We use the geometry learned with VLRC to lift dense vision-language features into 3D for zero-shot open-vocabulary semantic segmentation. As seen in Section~\ref{sec:method:problem_setup}, given a video sequence \(\{I_i\}_{i=1}^{n}\), the reconstruction model predicts depth, camera poses, and intrinsics, which induces a 3D point cloud in world coordinate of cardinality \(M\):

\[
\mathcal{P}=\{P_j\}_{j=1}^{M},
\qquad P_j\in\mathbb{R}^{3}.
\]

In parallel, a dense vision-language encoder \(\mathcal{E}^{2D}\) extracts a pixel-level feature map for each image:
\[
F_i=\mathcal{E}^{2D}(I_i),
\qquad
F_i(u,v)\in\mathbb{R}^{d}.
\]

For each 3D point \(P_j\) in world coordinate, we project it into every frame where it is visible. Let \(T_{\mathrm{w}\rightarrow i}\) denote the transformation from the world coordinate system to camera \(i\), and let \(K\) be the camera intrinsics. The projection of \(P_j\) into image \(I_i\) is given by
\[
\tilde q_{ij}
=
K\,T_{\mathrm{w}\rightarrow i}
\begin{bmatrix}
P_j\\
1
\end{bmatrix},
\qquad
q_{ij}
=
\pi(\tilde q_{ij})
=
(u_{ij},v_{ij}),
\]
 We then retrieve the corresponding dense VLM feature by bilinear sampling:
\[
f_{ij}=F_i(q_{ij}).
\]

The 3D feature associated with point \(P_j\) is obtained by aggregating its multi-view features:
\[
f^{3D}(P_j)
=
\frac{
\sum_{i=1}^{N}
\alpha_{ij}\,m_{ij}\,f_{ij}
}{
\sum_{i=1}^{N}
\alpha_{ij}\,m_{ij}
},
\]
where \(m_{ij}\in\{0,1\}\) indicates whether \(P_j\) is visible and validly projected in frame \(i\), and \(\alpha_{ij}\) is a view-dependent confidence weight.

Open-vocabulary segmentation can be performed in two modes. In the first mode, we assume a predefined set of semantic categories \(\mathcal{C}=\{c_k\}_{k=1}^{C}\). Given a VLM text encoder \(\mathcal{E}^{T}\), we compute one text embedding per class,
\[
t_k=\mathcal{E}^{T}(\mathrm{prompt}(c_k)),
\]
and assign each 3D point to the closest text embedding:
\[
\hat y_j
=
\arg\max_{k}
\cos\!\left(
f^{3D}(P_j),t_k
\right).
\]
This produces a zero-shot semantic label for every point in the predicted reconstruction.

In the second mode, the user provides an arbitrary free-form text query \(q\), such as \textit{``where is the Eifel-tower?''} in Fig.~\ref{fig:teaser_compact} or \textit{``Show me the entrance of the cathedral''} in Fig.~\ref{fig:sagrada}. We encode the query as
\[
t_q=\mathcal{E}^{T}(q),
\]
and compute a query relevance score for each 3D point:
\[
s_j(q)
=
\cos\!\left(
f^{3D}(P_j),t_q
\right).
\]
The resulting scores define a text-conditioned 3D activation heat map. For binary localization, we threshold the scores,
\[
\hat z_j(q)=\mathbf{1}\!\left[s_j(q)>\tau\right],
\]
where \(\tau\) is a similarity threshold. This produces a query-specific binary segmentation or localization mask in 3D.

Thus, the same fused 3D VLM representation supports both predefined-vocabulary semantic segmentation and free-form text-query localization.
For clarity, we present here the simple uniform averaging case, \(\alpha_{ij}=1\) for all \(i,j\), following~\cite{jiang2024open}. We also evaluate additional fusion strategies in Sec.~\ref{sec:experiments}.

\section{Experiments}
\label{sec:experiments}
Our experiments are organized around two questions:
(i)~Does VLRC improve core 3D estimates?
(ii)~Does the resulting geometry improve multi-view VLM feature fusion for open-vocabulary 3D segmentation?

\subsection{Implementation Details}
\label{sec:experiments:implementation_details}

We evaluate VLRC in two regimes: self-supervised SS3D fine-tuning and VGGT/SelfEvo-style adaptation. In both cases, VLRC is added as an auxiliary feature-space reprojection loss.

\paragraph{Self-supervised setting.}
For the self-supervised regime, we add VLRC to SS3D~\cite{hariat2026ss3d}, a monocular-video model trained with the photometric SfM objective in Eq.~\ref{eq:3d_loss_self} to predict depth, camera motion, and intrinsics. We start from the SS3D checkpoint pretrained on YouTube8M~\cite{abu2016youtube} and fine-tune it on the target datasets using the combined objective in Eq.~\ref{eq:final_loss}.

Following~\cite{hariat2023rebalancing}, we compute the validity mask \(m_{t,s}\) in Eq.~\ref{eq:vlrc_reprojection_loss} by comparing depth-pose reprojection correspondences with optical-flow correspondences. Pixels with large discrepancies, typically caused by dynamic objects, occlusions, or unreliable geometry, are excluded from the VLRC loss.

\paragraph{Supervised-pretrained setting.}
For the supervised-pretrained regime, we start from VGGT-1B~\cite{wang2025vggt} and follow the SelfEvo-style unlabeled post-training protocol~\cite{huang2026self}. A teacher processes the full sequence and provides stop-gradient reconstruction targets, while a reduced-context student is trained to match them. We add VLRC to this self-distillation objective. No pixel masking is used in this setting.

\paragraph{Dense VLM features.}
Unless otherwise stated, we use dense CLIP-derived vision-language features from~\cite{luddecke2022image}. The VLM encoder is frozen throughout training. Feature maps are upsampled to the image resolution using Feat\-Up~\cite{fu2024featup}, and VLRC is computed with cosine dissimilarity between reprojected source features and target-frame features. We provide ablations over the choice of VLM backbone and feature representation in Table~\ref{table:results_dino_clip}.

\paragraph{Optimization.}
During training, images are resized so that their shortest side is 518 pixels, followed by a \(518\times518\) crop. We use \(\lambda_{\mathrm{VLRC}}=0.01\) for SS3D fine-tuning and \(\lambda_{\mathrm{VLRC}}=0.1\) for VGGT/SelfEvo-style adaptation. Unless otherwise stated, both regimes are trained for 20 epochs using Adam~\cite{kingma2014adam} with \(\beta_1=0.99\) and \(\beta_2=0.999\). The learning rate follows a cosine decay schedule from \(10^{-5}\) to \(10^{-8}\). All experiments are implemented in PyTorch~\cite{paszke2019pytorch}.

\subsection{Datasets and Evaluation Protocols}
\label{sec:experiments:datasets}

We evaluate VLRC along two axes: core 3D estimation and open-vocabulary 3D semantic segmentation. For self-supervised 3D estimation, we compare SS3D+VLRC against SS3D on depth, camera motion, and intrinsics. Depth is evaluated on KITTI~\cite{geiger2013vision} and NYUv2~\cite{silberman2012indoor}; camera motion on Sintel~\cite{butler2012naturalistic} and TUM-RGBD~\cite{sturm2012benchmark}; and intrinsics on Sintel. For supervised-pretrained adaptation, we evaluate VGGT/SelfEvo+VLRC on depth using KITTI and Sintel. For open-vocabulary 3D segmentation, we evaluate on ScanNet200~\cite{dai2017scannet} and introduce a KITTI-based zero-shot protocol, described in Appendix~\ref{sec:appendix:segmentation_protocol}.

\subsection{Depth Estimation}
\label{sec:experiments:depth_results}

Tables~\ref{table:results_depth_kitti_ft} and~\ref{table:results_depth_nyu_ft} show that VLRC improves SS3D depth estimation on both KITTI and NYUv2, outperforming prior self-supervised baselines. Table~\ref{table:results_selfevo_depth} shows that VLRC also improves SelfEvo-style adaptation of VGGT on Sintel and KITTI, indicating that the signal complements both photometric self-supervision and supervised-pretrained unlabeled adaptation. More results on camera motion and intrinsics are given in Appendix~\ref{sec:appendix:more_results}.

Table~\ref{table:results_dino_clip} studies the feature backbone used in VLRC when fine-tuning SS3D + VLRC on KITTI. CLIP-Seg outperforms DINOv2 and slightly improves over TIPSv2. While TIPSv2 is designed to improve dense patch-text alignment through spatially aware vision-language pretraining, CLIP-Seg builds on CLIP with an explicit dense segmentation decoder.

\begin{table*}[t]
\centering
\setlength{\tabcolsep}{2.2pt}
\renewcommand{\arraystretch}{1.05}
\scriptsize

\begin{minipage}[t]{0.49\textwidth}

\resizebox{\linewidth}{!}{%
\begin{tabular}{l c c c c c c c c}
\toprule
\multirow{2}{*}{\textbf{Method}} 
& \multirow{2}{*}{\textbf{Self-Supervised}}
& \multicolumn{4}{c}{\textit{Lower is better} $\downarrow$} 
& \multicolumn{3}{c}{\textit{Higher is better} $\uparrow$} \\
\cmidrule(r){3-6} \cmidrule(l){7-9}
& 
& \textcolor{lightblue}{Abs Rel} 
& Sq Rel 
& RMSE 
& \textcolor{lightblue}{RMSE log}
& \textcolor{lightblue}{$\delta_{1}$} 
& $\delta_{2}$ 
& $\delta_{3}$ \\
\midrule

Monodepth2~\cite{godard2019digging}              & \cmark & 0.110 & 0.831 & 4.642 & 0.187 & 0.883 & 0.962 & 0.982 \\
MonoViT~\cite{zhao2022monovit}                   & \cmark & 0.099 & 0.708 & 4.372 & 0.175 & 0.900 & 0.967 & 0.984 \\
HR-Depth~\cite{lyu2021hr}                        & \cmark & 0.109 & 0.792 & 4.632 & 0.185 & 0.884 & 0.962 & 0.983 \\
RA-Depth~\cite{he2022ra}                         & \cmark & 0.096 & 0.613 & 4.216 & 0.171 & 0.903 & 0.968 & 0.985 \\
DIFFNet~\cite{zhou_diffnet}                      & \cmark & 0.102 & 0.764 & 4.483 & 0.180 & 0.896 & 0.965 & 0.983 \\
Hariat \textit{et al.}~\cite{hariat2025improved} & \cmark & 0.082 & 0.604 & 4.108 & 0.162 & 0.928 & 0.968 & 0.985 \\
\midrule
SS3D                                    & \cmark & 0.064 & 0.530 & 3.212 & 0.138 & 0.946 & 0.977 & 0.986 \\
\textbf{Ours: SS3D + VLRC}                                    & \cmark & \textbf{0.060} & \textbf{0.496} & \textbf{2.908} & \textbf{0.133} & \textbf{0.950} & \textbf{0.978} & \textbf{0.986} \\

\bottomrule
\end{tabular}%
}
\caption{\textbf{KITTI FT} - Fine-tuning on KITTI~\cite{geiger2013vision} and evaluating on KITTI. }
\label{table:results_depth_kitti_ft}

\end{minipage}
\hfill
\begin{minipage}[t]{0.49\textwidth}

\resizebox{\linewidth}{!}{%
\begin{tabular}{l c c c c c c c}
\toprule
\multirow{2}{*}{\textbf{Method}} 
& \multirow{2}{*}{\textbf{Self-Supervised}}
& \multicolumn{3}{c}{\textit{Lower is better} $\downarrow$} 
& \multicolumn{3}{c}{\textit{Higher is better} $\uparrow$} \\
\cmidrule(r){3-5} \cmidrule(l){6-8}
& 
& \textcolor{lightblue}{Abs Rel} 
& RMSE
& \textcolor{lightblue}{RMSE log}
& \textcolor{lightblue}{$\delta_{1}$} 
& $\delta_{2}$ 
& $\delta_{3}$ \\
\midrule

MovingIndoor~\cite{zhou2019moving}                    & \cmark & 0.208 & 0.712 & 0.086 & 0.674 & 0.900 & 0.968 \\
StructDepth~\cite{li2021structdepth}                  & \cmark & 0.140 & 0.540 & 0.060 & 0.817 & 0.955 & 0.988 \\
MonoIndoor++~\cite{li2022monoindoor++}                & \cmark & 0.132 & 0.517 & \texttt{N/A} & 0.834 & 0.961 & 0.990 \\
IndoorDepth~\cite{fan2023deeper}                      & \cmark & 0.126 & 0.494 & 0.054 & 0.845 & 0.965 & 0.991 \\
Hariat \textit{et al.}~\cite{hariat2025improved}      & \cmark & 0.115 & 0.458 & 0.054 & 0.859 & 0.970 & 0.992 \\
\midrule
SS3D                                       & \cmark & 0.090 & 0.418 & 0.049 & 0.866 & 0.970 & 0.992 \\
\textbf{Ours: SS3D + VLRC}                                         & \cmark & \textbf{0.082} & \textbf{0.407} & \textbf{0.044} & \textbf{0.867} & \textbf{0.971} & \textbf{0.992} \\
\bottomrule
\end{tabular}%
}
\caption{\textbf{NYU FT} - Fine-tuning on NYUv2~\cite{silberman2012indoor} and evaluating on NYUv2.}
\label{table:results_depth_nyu_ft}

\end{minipage}
\end{table*}

\begin{table}[t]
\centering
\begin{tabular}{lcccc}
\toprule
\textbf{Method} 
& \textbf{AbsRel $\downarrow$} 
& \textbf{$\Delta$ AbsRel} 
& \textbf{$\delta_1$ (\%) $\uparrow$} 
& \textbf{$\Delta\,\delta_1$} \\
\midrule
SS3D (Baseline)
& 0.064
& --
& 94.6
& -- \\

w/ DINOv2~\cite{oquab2023dinov2}
& 0.065
& \textcolor{red!70}{+0.001}
& 94.3
& \textcolor{red!70}{-0.3} \\

w/ TIPSv2~\cite{cao2026tipsv2}
& 0.061
& \textcolor{green!60}{-0.003}
& 95.0
& \textcolor{green!60}{+0.3} \\

w/ CLIP-Seg~\cite{luddecke2022image}
& \textbf{0.060}
& \textcolor{green!60}{\textbf{-0.004}}
& \textbf{95.0}
& \textcolor{green!60}{\textbf{+0.4}} \\
\bottomrule
\end{tabular}
\caption{
\textbf{Effect of the feature backbone used in VLRC.} Fine-tuning and evaluation are performed on KITTI.
\(\Delta\) values are computed relative to SS3D. Green indicates improvement and red indicates degradation.
}
\label{table:results_dino_clip}
\end{table}

\subsection{Open-Vocabulary Semantic Segmentation}
\label{sec:experiments:semantic_results}

Most existing open-vocabulary 3D semantic segmentation protocols rely on ground-truth geometry, making it difficult to evaluate whether a reconstruction model produces geometry that is well aligned with dense VLM features. We therefore evaluate VLRC under two complementary settings.


For the ScanNet200 protocol, we use Casper3D~\cite{hariat2026lightweight} as the downstream 3D semantic representation model. Casper3D fuses multi-view 2D features into a view-invariant 3D representation. To test the effect of VLRC on geometry-aware feature fusion, we pretrain Casper3D on NYUv2, a dataset visually and geometrically close to ScanNet, using reconstructions from SS3D, SS3D+VLRC, SelfEvo, and SelfEvo+VLRC. We then fine-tune Casper3D on ScanNet following the original protocol. This evaluates whether VLRC-improved geometry leads to stronger downstream 3D semantic representations.

\begin{figure}[!t]
    \centering
    \setlength{\tabcolsep}{2pt}
    \renewcommand{\arraystretch}{0.85}

    \begin{tabular}{cc}
        \small \textbf{Without VLRC} & \small \textbf{With VLRC} \\[-1pt]

        \includegraphics[width=0.48\linewidth]{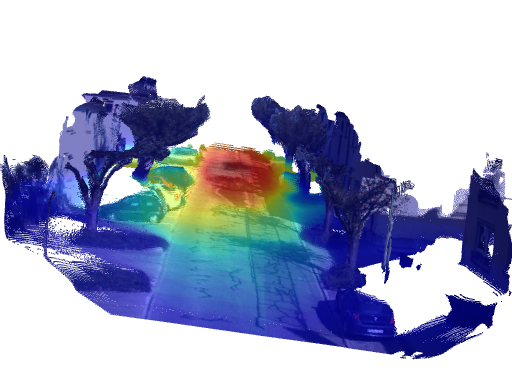} &
        \includegraphics[width=0.48\linewidth]{images/supp/qualitative/kitti_2_heat_map.png} \\[-2pt]
        \multicolumn{2}{c}{\small (a) Prompt: \textit{``Where are the cars?''}} \\[4pt]

        \includegraphics[width=0.48\linewidth]{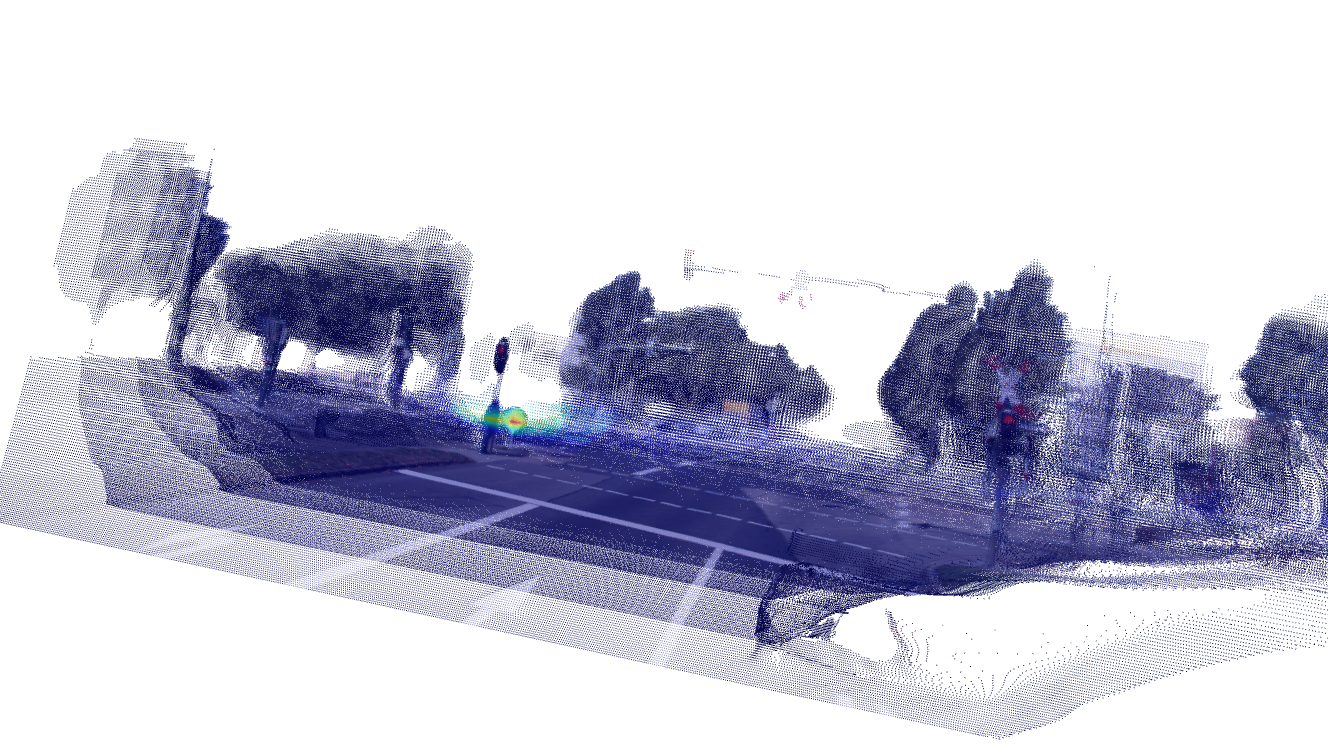} &
        \includegraphics[width=0.48\linewidth]{images/supp/qualitative/kitti_3_heat_map.png} \\[-2pt]
        \multicolumn{2}{c}{\small (b) Prompt: \textit{``Where is the person?''}} \\
    \end{tabular}

\caption{
\textbf{Effect of VLRC on open-vocabulary 3D localization.}
Without VLRC, cars are not well located and the person is missed. Adding VLRC produces cleaner and more spatially coherent 3D activations.
}
    \label{fig:vlrc_failure_cases}
\end{figure}

Second, many 3D estimation methods provide checkpoints or reported results on KITTI. We therefore introduce a KITTI-based zero-shot open-vocabulary 3D semantic segmentation protocol. This setting is designed to compare reconstruction models under an identical evaluation pipeline. Each method first predicts a 3D point cloud. Each 3D point is then projected into a fixed number of adjacent frames, where dense CLIP logits are extracted and aggregated across views. We compute CLIP logits using Cityscapes semantic categories as text prompts, matching the label space of the SegFormer pseudo labels. Further details are provided in Appendix~\ref{sec:appendix:segmentation_protocol}. This protocol directly evaluates the alignment between the predicted geometry and dense CLIP features. Fig.~\ref{fig:vlrc_failure_cases} illustrates typical failure cases when VLRC is not used. Without VLRC, the predicted geometry induces less reliable cross-view correspondences, causing dense VLM activations to be fused at inconsistent 3D locations. This produces noisy, fragmented, or misplaced prompt responses in the reconstructed scene. In contrast, adding VLRC encourages the geometry to preserve feature-level consistency across views, resulting in cleaner and more spatially coherent open-vocabulary 3D localization. 

Results are shown in Table~\ref{table:results_semantic}. VLRC improves ScanNet200 performance under the Casper3D protocol and substantially improves KITTI zero-shot segmentation, indicating better alignment between predicted 3D structure and dense vision-language features.

\begin{figure}[t]
    \centering
    \setlength{\tabcolsep}{3pt}
    \renewcommand{\arraystretch}{0.05}

    \begin{tabular}{c c}
        \includegraphics[height=2.0cm,width=0.42\linewidth,keepaspectratio]{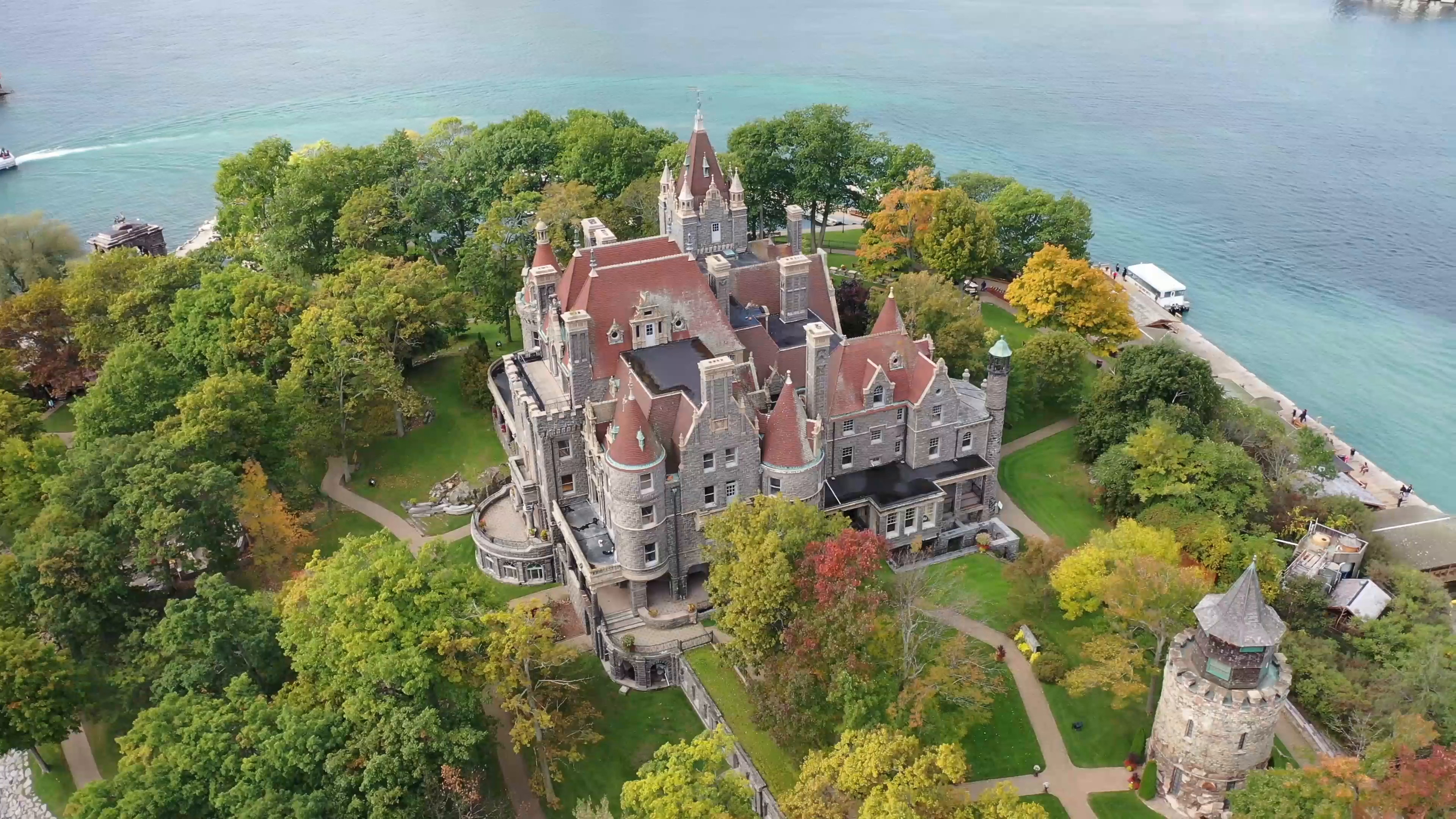} &
        \includegraphics[height=2.0cm,width=0.42\linewidth,keepaspectratio]{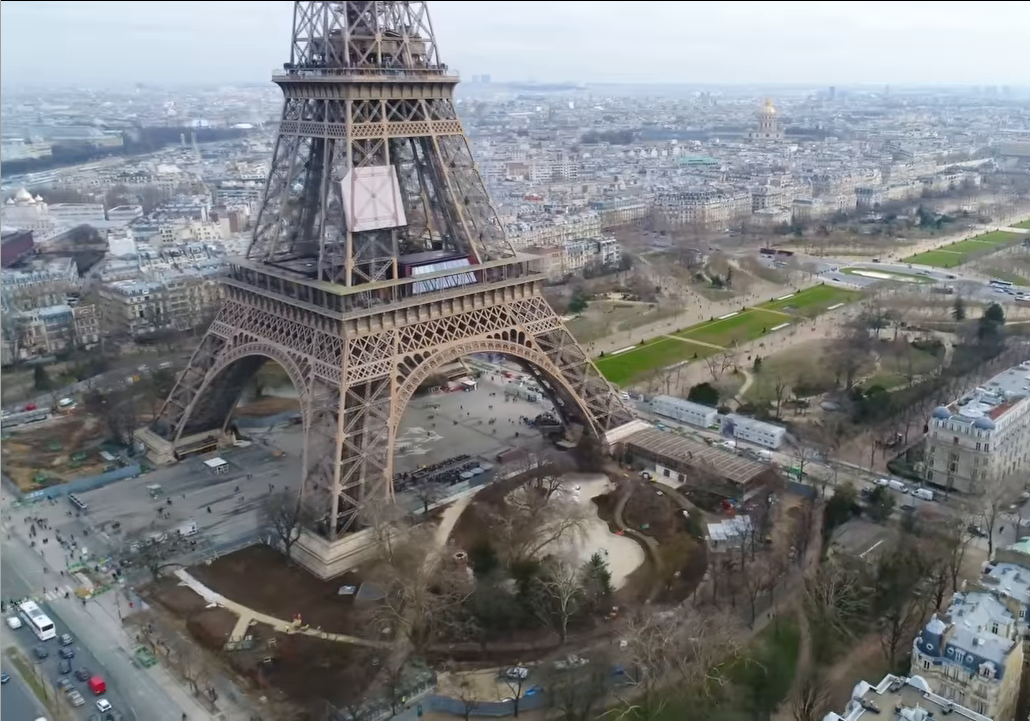} \\

        \includegraphics[height=2.0cm,width=0.42\linewidth,keepaspectratio]{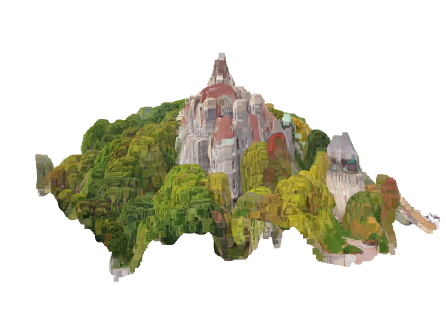} &
        \includegraphics[height=2.0cm,width=0.42\linewidth,keepaspectratio]{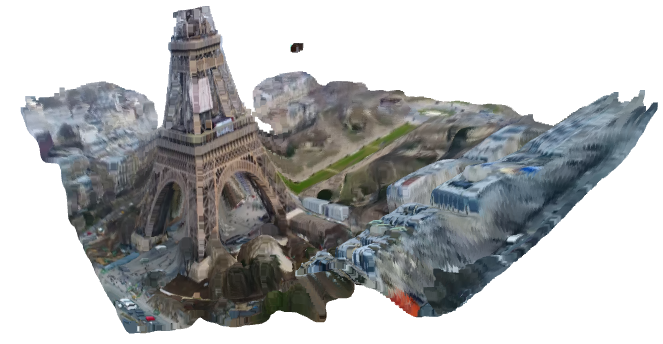} \\

        \includegraphics[height=2.0cm,width=0.42\linewidth,keepaspectratio]{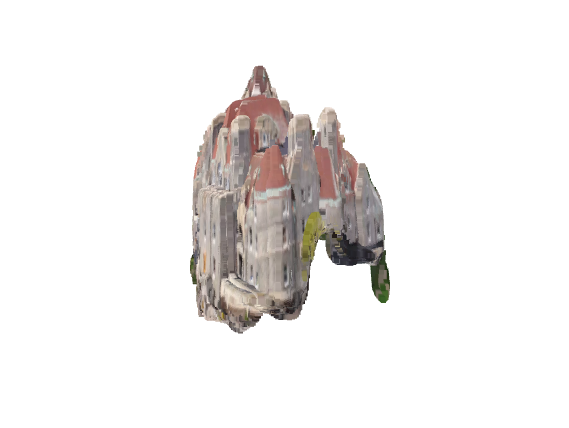} &
        \includegraphics[height=2.0cm,width=0.42\linewidth,keepaspectratio]{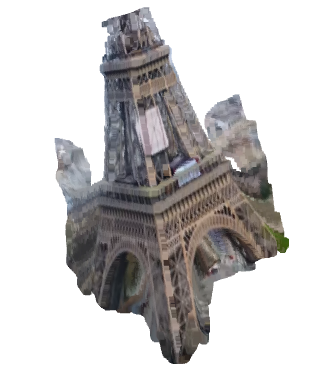} \\

    \end{tabular}

    \vspace{-6pt}
    \caption{
    \textbf{Zero-shot open-vocabulary 3D semantic localization from casual monocular videos.}
    For each scene, we show Top: one input frame of the video, Middle: 3D reconstruction from SS3D + VLRC fine-tuned on web-videos, Bottom: the CLIP-based 3D localization from text prompts that are respectively \textit{``Where is the Castle?''} and \textit{``Where is the Eiffel Tower?''}. For CLIP-based 3D localization, we retain only 3D points with text similarity above a threshold of \(\tau=0.5\).
    }
    \label{fig:teaser_compact}
\end{figure}

\begin{figure*}[t]
    \centering
    \setlength{\tabcolsep}{2pt}
    \renewcommand{\arraystretch}{0.15}

    \begin{tabular}{cccc}
        \includegraphics[width=0.245\linewidth,height=2.5cm]{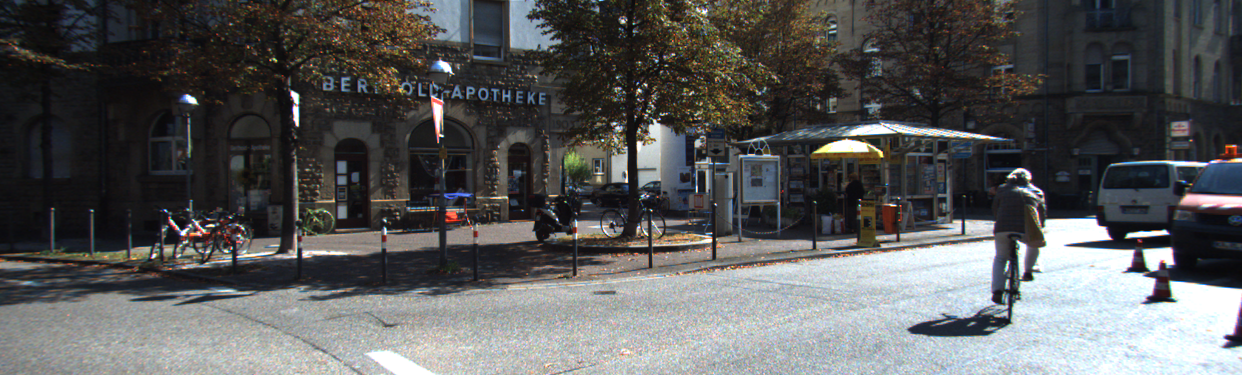} &
        \includegraphics[width=0.245\linewidth,height=2.5cm]{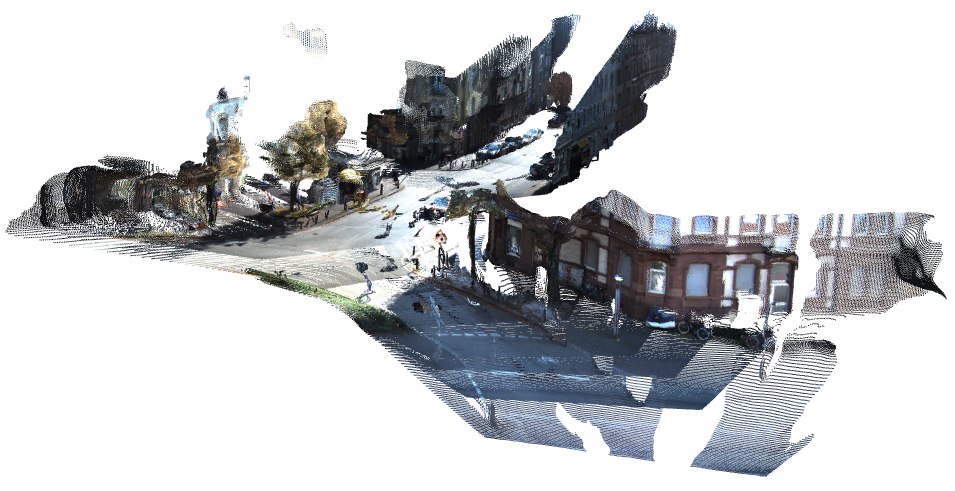} &
        \includegraphics[width=0.245\linewidth,height=2.5cm]{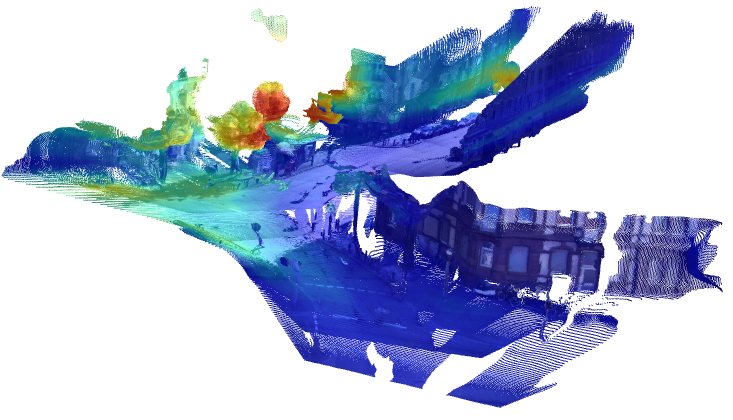} &
        \includegraphics[width=0.245\linewidth,height=2.5cm]{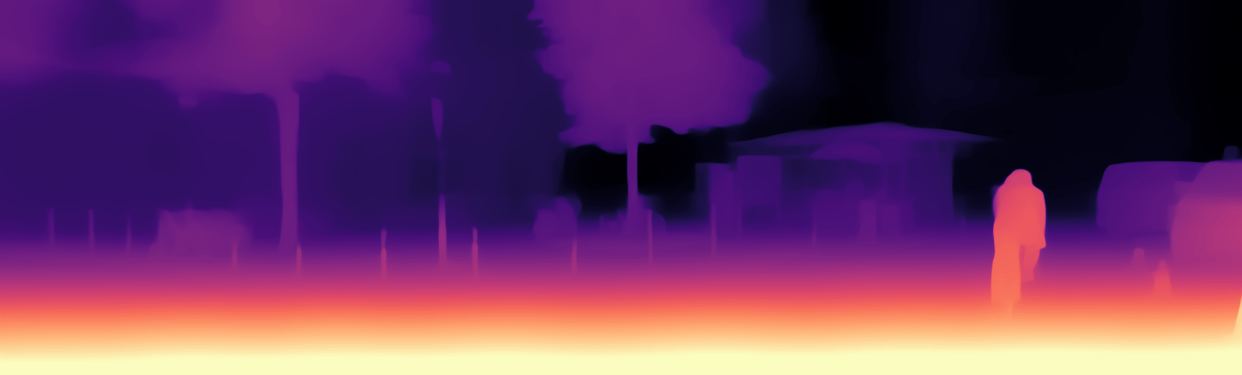} \\[-2pt]

        \includegraphics[width=0.245\linewidth,height=2.5cm]{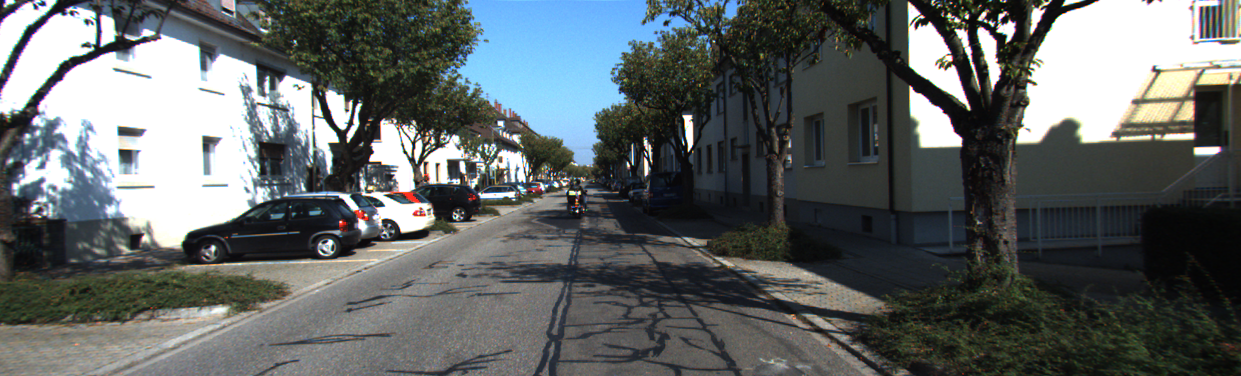} &
        \includegraphics[width=0.245\linewidth,height=2.5cm]{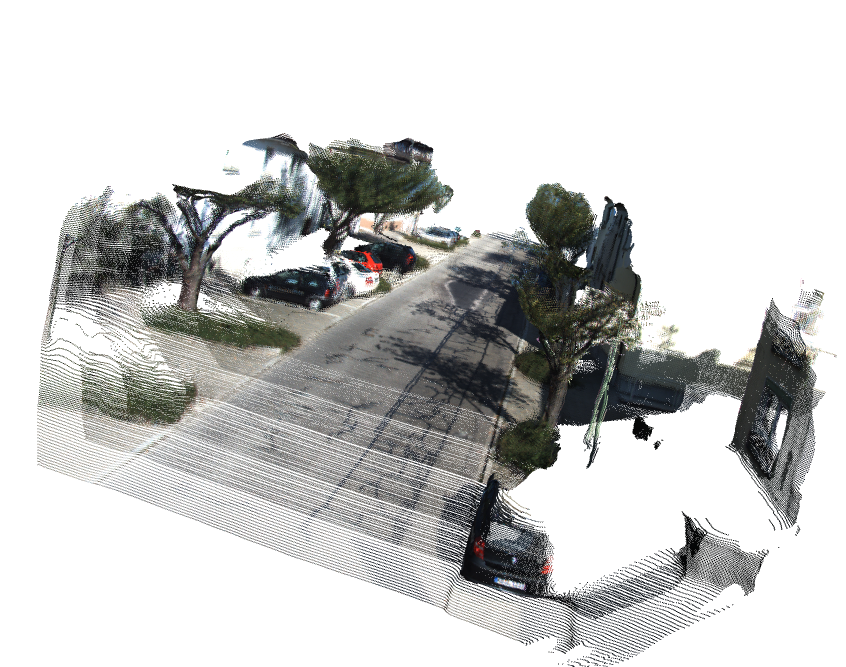} &
        \includegraphics[width=0.245\linewidth,height=2.5cm]{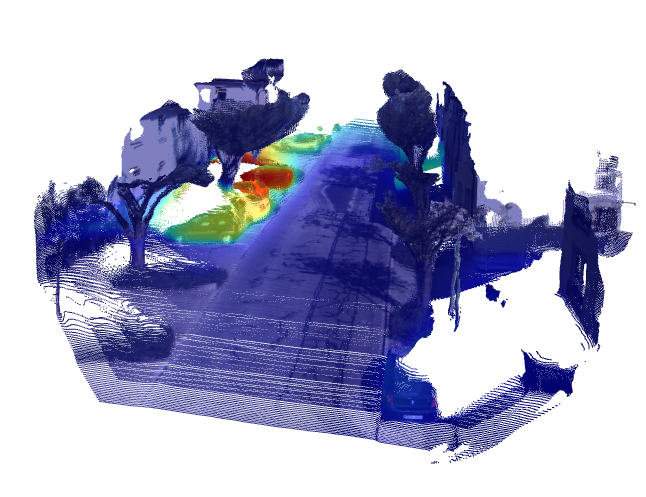} &
        \includegraphics[width=0.245\linewidth,height=2.5cm]{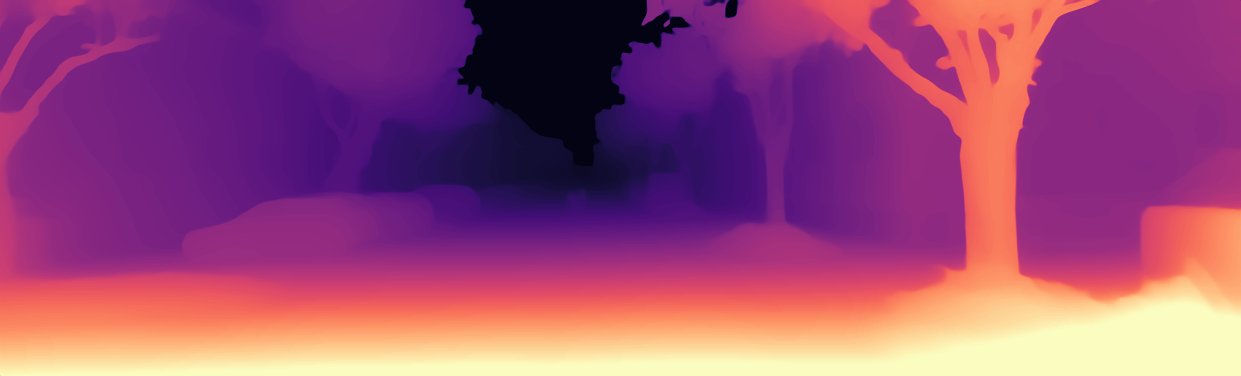} \\[-2pt]

        \includegraphics[width=0.245\linewidth,height=2.5cm]{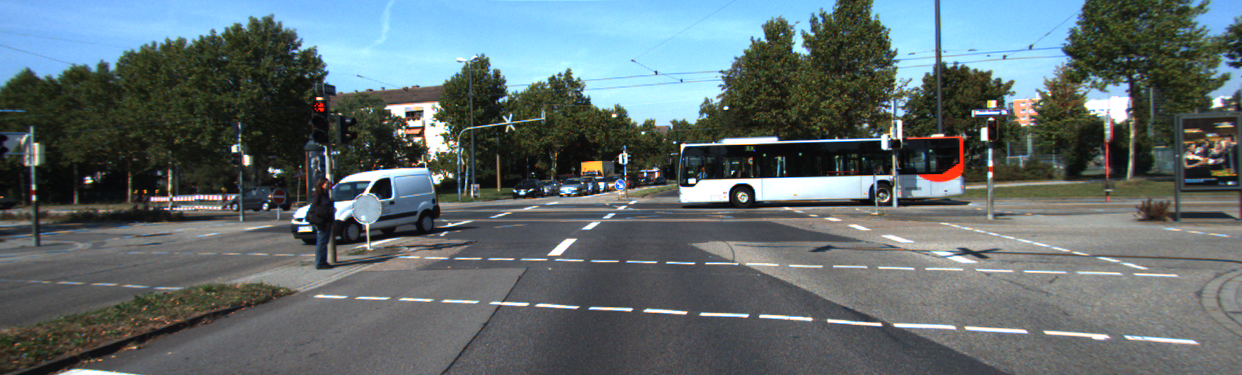} &
        \includegraphics[width=0.245\linewidth,height=2.5cm]{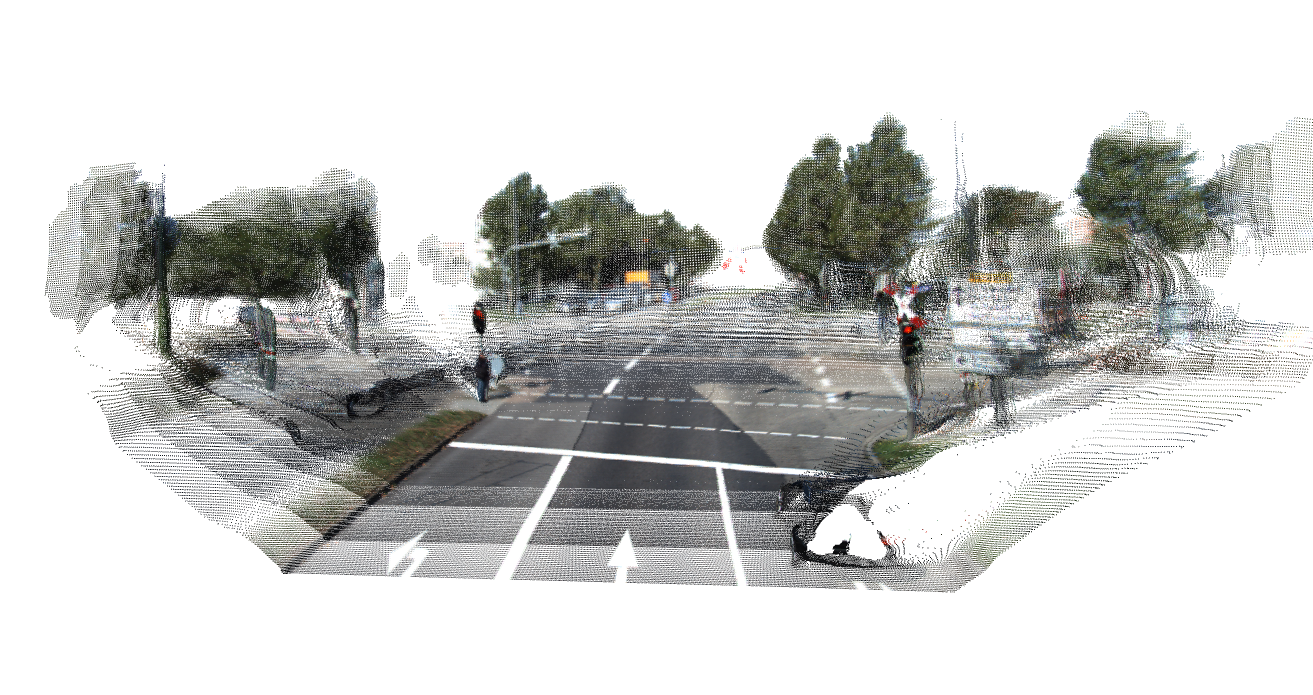} &
        \includegraphics[width=0.245\linewidth,height=2.5cm]{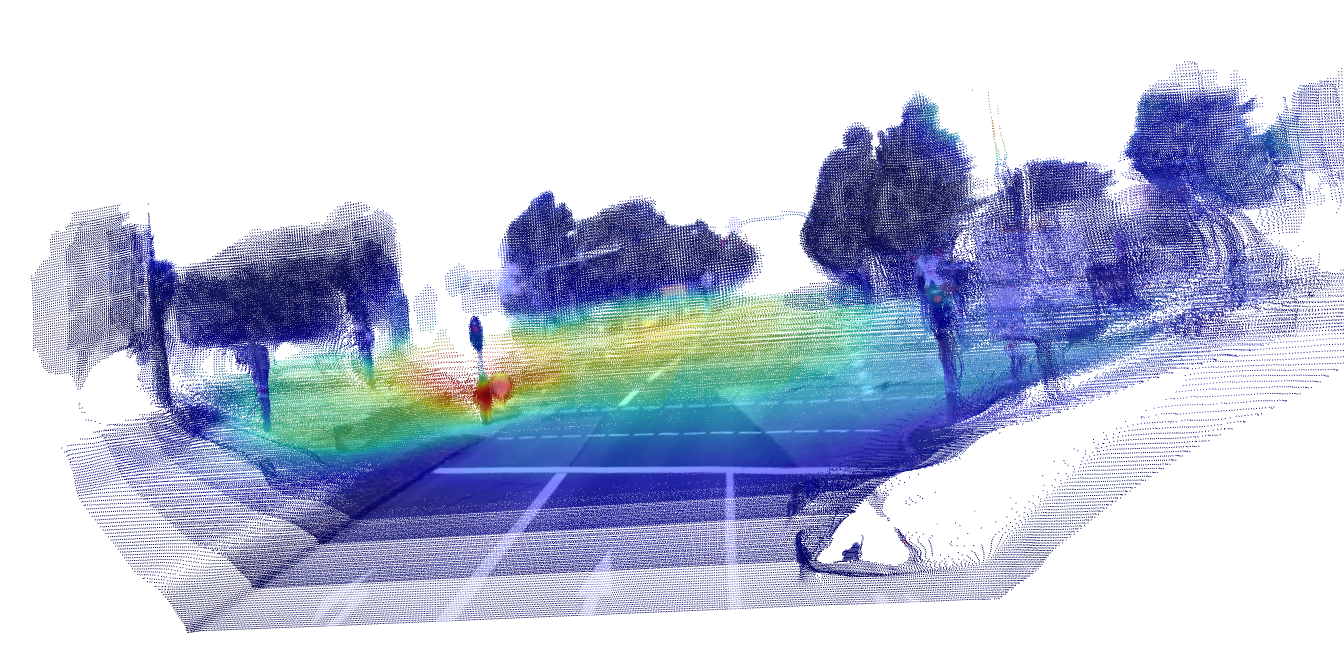} &
        \includegraphics[width=0.245\linewidth,height=2.5cm]{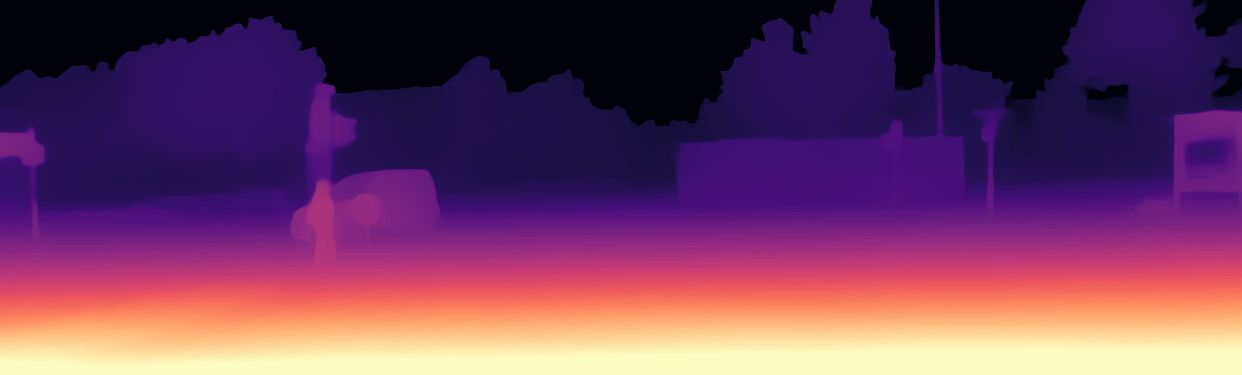}
    \end{tabular}

    \vspace{-4pt}
    \caption{
        Qualitative open-vocabulary 3D results on KITTI. Each row shows, from left to right:
        \textbf{Input frame}, \textbf{Point cloud}, \textbf{Prompt heatmap}, and \textbf{Depth}.
        The heatmaps are generated from text queries that are respectively from top to bottom:
        \textit{``Where are the trees?''}, \textit{``Where are the cars?''}, and
        \textit{``Where is the person?''}.
    }
    \label{fig:qualitative_kitti}
\end{figure*}

Qualitative results for arbitrary free-form text queries are shown in Fig.~\ref{fig:qualitative_kitti}, Fig.~\ref{fig:sagrada}, and Fig.~\ref{fig:teaser_compact}. Fig.~\ref{fig:qualitative_kitti} uses SS3D+VLRC fine-tuned on KITTI, while Fig.~\ref{fig:sagrada} and Fig.~\ref{fig:teaser_compact} show SS3D+VLRC trained on YouTube8M web videos.

\begin{table}[!t]
\centering
\resizebox{\linewidth}{!}{
\begin{tabular}{lcccc}
\toprule
\multirow{2}{*}{Method}
& \multicolumn{2}{c}{Sintel}
& \multicolumn{2}{c}{KITTI} \\
\cmidrule(lr){2-3}
\cmidrule(lr){4-5}
& Abs Rel $\downarrow$ & $\delta < 1.25 \uparrow$
& Abs Rel $\downarrow$ & $\delta < 1.25 \uparrow$ \\
\midrule
VGGT~\cite{wang2025vggt}
& 0.227 & 0.684
& 0.059 & 0.961 \\
SelfEvo (VGGT)
& 0.212 & 0.692
& 0.042 & 0.979 \\
\textbf{SelfEvo + VLRC}
& \textbf{0.209} & \textbf{0.700}
& \textbf{0.038} & \textbf{0.980} \\
\bottomrule
\end{tabular}
}
\caption{
\textbf{VLRC complements SelfEvo-style post-training.}
Adding VLRC to SelfEvo improves depth prediction on Sintel and KITTI, showing that VLRC provides complementary supervision during unlabeled adaptation.
}
\label{table:results_selfevo_depth}
\end{table}

\begin{table}[!t]
\centering
\resizebox{\linewidth}{!}{
\begin{tabular}{lcccc}
\toprule
\multirow{2}{*}{Method}
& \multicolumn{2}{c}{ScanNet200}
& \multicolumn{2}{c}{KITTI} \\
\cmidrule(lr){2-3}
\cmidrule(lr){4-5}
& mIoU $\uparrow$ & mAcc $\uparrow$
& mIoU $\uparrow$ & mAcc $\uparrow$ \\
\midrule
\multicolumn{5}{c}{\textit{ScanNet200: Casper3D protocol}} \\
\midrule
Casper3D~\cite{hariat2026lightweight}
& 11.0 & 18.1 & -- & -- \\
Casper3D pretrained w/ SelfEvo geometry
& 11.1 & 18.3 & -- & -- \\
Casper3D pretrained w/ SelfEvo+VLRC geometry
& \textbf{12.1} & \textbf{19.0} & -- & -- \\
\midrule
\multicolumn{5}{c}{\textit{KITTI: zero-shot geometry-to-semantics protocol}} \\
\midrule
SS3D
& -- & -- & 17.5 & 28.3 \\
SS3D + VLRC
& -- & -- & \textbf{24.0} & \textbf{39.3} \\
\bottomrule
\end{tabular}
}
\caption{
\textbf{Open-vocabulary 3D semantic segmentation.}
We evaluate VLRC under two complementary protocols: ScanNet200 with Casper3D fine-tuning, and a KITTI zero-shot geometry-to-semantics protocol. VLRC improves performance in both settings, indicating better alignment between predicted 3D structure and dense vision-language features.
}
\label{table:results_semantic}
\end{table}

\section{Conclusion}
\label{sec:conclusion}

We introduced \textbf{Vision-Language Reprojection Consistency}, a general auxiliary objective for feed-forward 3D learning. Instead of using vision-language models only after geometry has been estimated, VLRC uses dense VLM features as a training signal: predicted depth, camera motion, and intrinsics induce cross-view correspondences, and the model is encouraged to make language-aligned features consistent across these reprojected views. This provides a scalable feature-space supervision signal that does not require additional 3D annotations and complements both photometric self-supervision and supervised-pretrained 3D reconstruction models.

Across self-supervised SS3D fine-tuning and VGGT/SelfEvo-style unlabeled adaptation, VLRC improves core 3D estimates . We further show that geometry trained with VLRC better supports multi-view aggregation of dense VLM features, improving open-vocabulary 3D semantic segmentation on both indoor and outdoor protocols. These results suggest that aligning geometry with vision-language representations during training is a promising direction for building scalable 3D models that are geometrically accurate, and more compatible with open-vocabulary semantic understanding.

\paragraph{Future work.}
An interesting direction for future work is to jointly adapt the VLM together with the 3D reconstruction model, so that the dense vision-language features themselves become more geometry-aware.

{
    \small
    \bibliographystyle{ieeenat_fullname}
    \bibliography{main}
}

\appendix
\clearpage
\setcounter{page}{1}
\setcounter{section}{0}
\maketitlesupplementary
\renewcommand\thesection{\Alph{section}}
\newtheorem{mytheorem}{Theorem}
\setcounter{figure}{0}

\numberwithin{figure}{section}
\numberwithin{equation}{section}
\numberwithin{table}{section}

\noindent
The supplementary material includes multiple details and insights that complement the main paper.

\section{KITTI Open-Vocabulary 3D Segmentation Protocol}
\label{sec:appendix:segmentation_protocol}
Our new protocol is based on the sequences of the KITTI Odometry dataset
To obtain the 3D segmentation map, for each scene we first consider a temporal window of 10 frames centered around the target frame. Using the predicted depth, intrinsic parameters, and camera poses, of the considered pretrained feed-forward multi-view reconstruction model  we back-project the target frame into a 3D point cloud. Each 3D point is then reprojected into all 10 neighboring frames, and we aggregate the corresponding logits with respect to  each class with the prompt template  \texttt{``a photo of a [CLASS].''} and by averaging them across views. The final semantic label for each 3D point is obtained by taking the 
$\arg\max$ over the averaged logits.\\
For evaluation, we use Velodyne LiDAR points as reference 3D points. Since KITTI Odometry does not provide dense semantic annotations, we assign pseudo-semantic labels by applying SegFormer-B5~\cite{xie2021segformer} to the RGB frames and projecting the LiDAR points into the labeled images. 

To facilitate reproducibility, we provide the code for generating the pseudo labels, constructing the 3D point-level evaluation set, and running the KITTI open-vocabulary 3D segmentation protocol.

\begin{figure*}[t]
    \centering
    \setlength{\tabcolsep}{2pt}
    \renewcommand{\arraystretch}{0.2}

    \begin{tabular}{cccc}
        \includegraphics[width=0.18\linewidth]{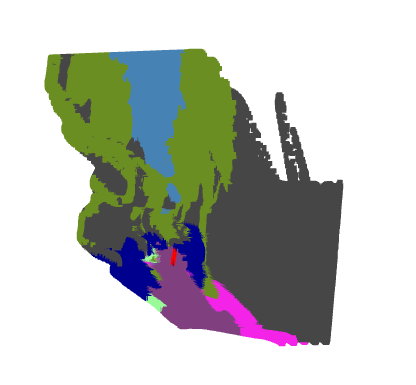} &
        \includegraphics[width=0.18\linewidth]{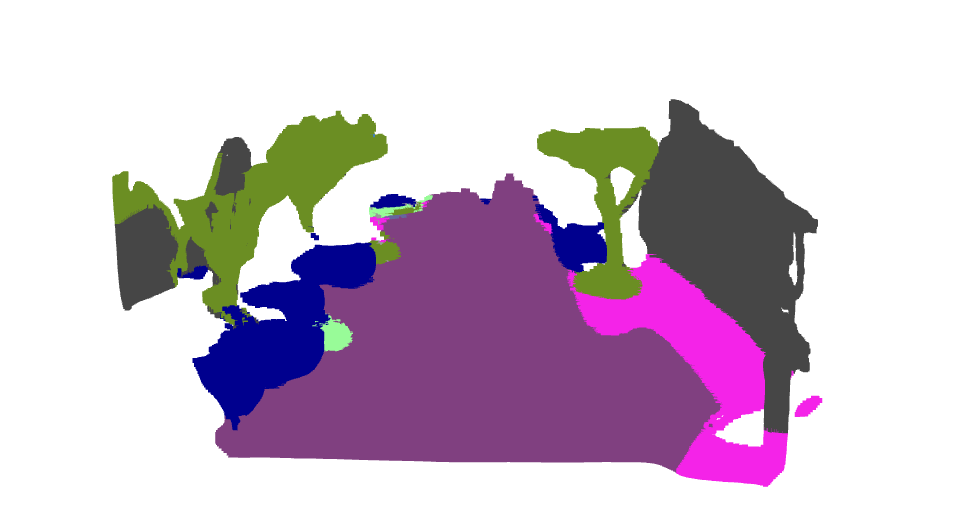} &
        \includegraphics[width=0.18\linewidth]{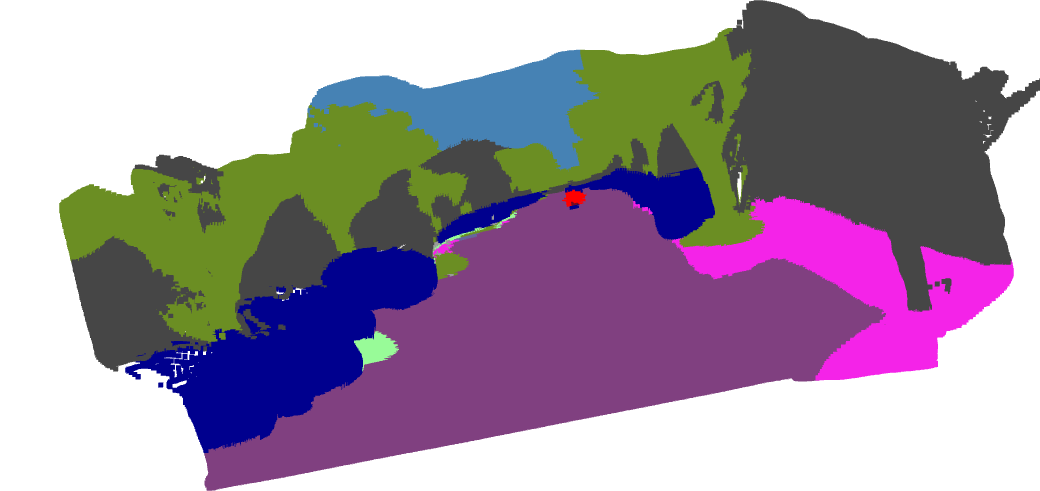} &
        \includegraphics[width=0.18\linewidth]{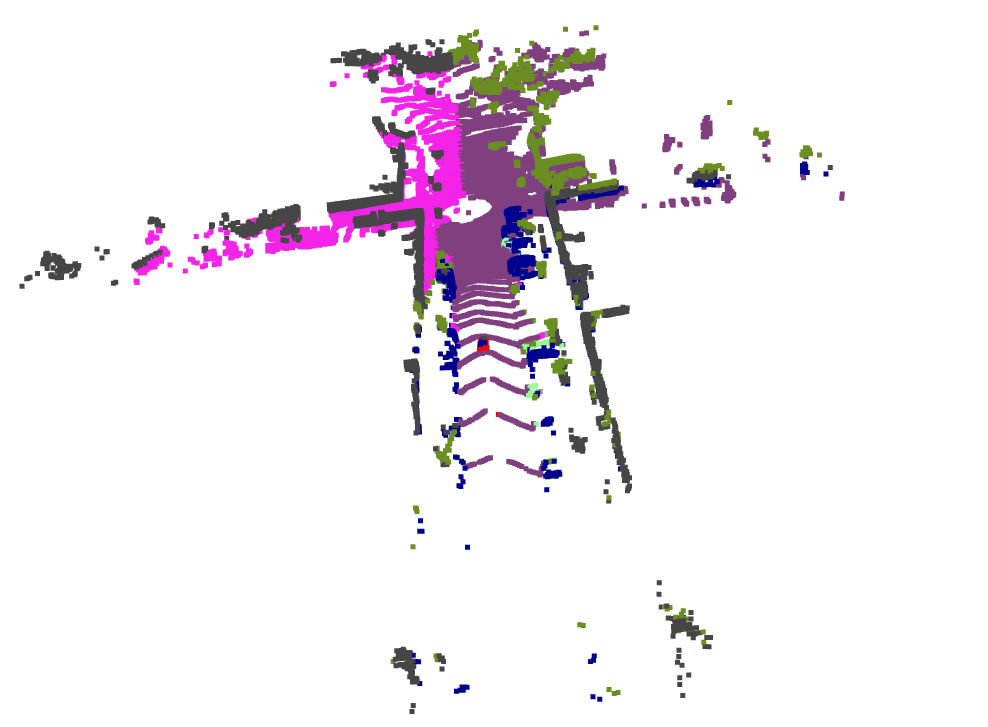} \\[-2pt]
        \small AnyCam & \small Ours & \small DUSt3R & \small KITTI GT \\[-1pt]
    \end{tabular}

    \vspace{2pt}

    \includegraphics[width=0.58\linewidth]{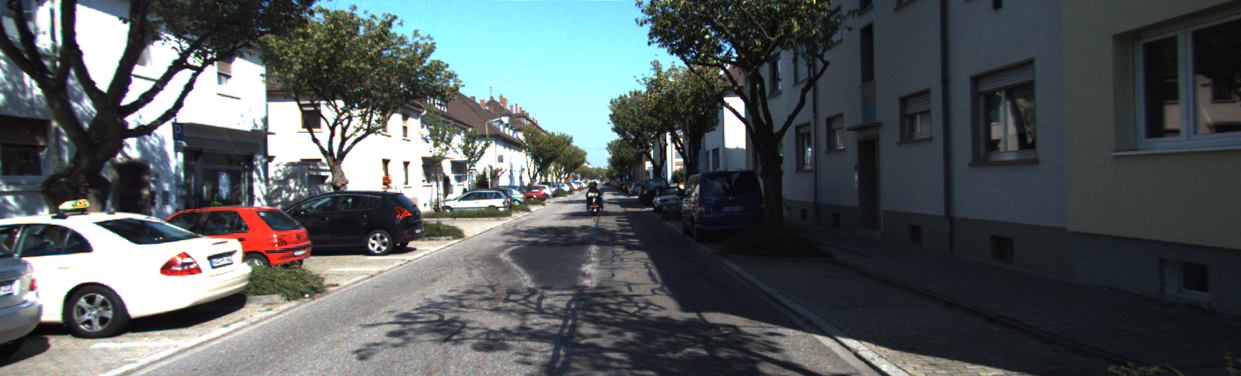}

    \vspace{-4pt}
    \caption{
        Comparison of labeled point clouds obtained by reprojection from \textbf{AnyCam\cite{wimbauer2025anycam}}, \textbf{SS3D + VLRC}, and \textbf{DUSt3R}, with KITTI ground truth shown for reference. For visualization, reprojections are computed from ground-truth labels rather than CLIP predictions. AnyCam exhibits strong geometric distortions, while DUSt3R benefits from supervised training and heavier test-time alignment/bundle adjustment. In contrast, our method remains fully self-supervised and lightweight at inference.
    }
    \label{fig:reprojections_comparison}
\end{figure*}

This protocol directly evaluates the alignment between the predicted geometry and dense CLIP features. As shown in Tab.~\ref{tab:3d_semseg}, SS3D + VLRC outperforms AnyCam\cite{wimbauer2025anycam} and VGGT\cite{wang2025vggt}, and approaches the performance of DUSt3R\cite{wang2024dust3r}, which are supervised methods. In contrast, SS3D+VLRC is trained fully self-supervised. DUSt3R remains a strong reference because on top of being supervised, it relies on heavier test-time optimization. Qualitative comparisons are shown on Fig.~\ref{fig:reprojections_comparison}.\\

We also observe that computing CLIP logits over SAM masks, rather than individual pixels, provides an additional improvement.

\begin{table}[t]
    \centering
    \small
    \renewcommand{\arraystretch}{1.15}
    \begin{tabular}{lcc}
        \toprule
        \textbf{Method} & \textbf{mIoU (\%)$\uparrow$} & \textbf{mAcc (\%)$\uparrow$} \\
        \midrule
        DUSt3R                                & 26.2 & 43.4 \\
        AnyCam                                & 15.2 & 24.0 \\
        VGGT                                  & 18.2 & 32.2 \\
        \midrule
        SS3D  & 17.5 & 28.3 \\
        Ours                                   & 24.0 & 39.3 \\
        \bottomrule
    \end{tabular}
    \caption{\textbf{3D semantic segmentation results.}
    Mean IoU (mIoU) and mean class accuracy (mAcc).}
    \label{tab:3d_semseg}
\end{table}

\section{Additional 3D Estimation Results}
\label{sec:appendix:more_results}
For pose and intrinsic results, we retrain SS3D + VLRC on Youtube8M\cite{abu2016youtube} and we evaluate the performances in zero-shot for camera motion on Sintel~\cite{butler2012naturalistic} and TUM-RGBD~\cite{sturm2012benchmark}; and intrinsics on Sintel.
 
\paragraph{Pose Estimation.}
We further report zero-shot pose estimation results in Table~\ref{table:results_pose}. We evaluate on Sintel, which contains synthetic scenes with complex motion and strong appearance changes, and on dynamic TUM-RGBD, which captures challenging real-world motion. We compare against AnyCam, a recent camera-pose method designed to leverage strong pretrained supervised depth and flow estimators. Despite being trained fully self-supervised, SS3D+VLRC achieves competitive or stronger performance across these benchmarks. The gains over SS3D show that the VRLC signal improves not only depth, but also the camera-motion component of the unified 3D estimator.

\begin{table*}[!t]
\centering
\small
\begin{tabular}{l l c c ccc ccc}
\toprule
Category & Method & No Superv. & Approx. Runtime &
\multicolumn{3}{c}{Sintel} &
\multicolumn{3}{c}{TUM-RGBD (dynamics)} \\
 & & & & ATE$\downarrow$ & RPE$_{\text{trans}}\downarrow$ & RPE$_{\text{rot}}\downarrow$ &
ATE$\downarrow$ & RPE$_{\text{trans}}\downarrow$ & RPE$_{\text{rot}}\downarrow$ \\
\midrule

& AnyCam\cite{wimbauer2025anycam} & \checkmark &  $<20$sec & 0.099 & 0.045 & \textbf{0.567} & 0.095 & 0.025 & 1.050 \\
& SS3D\cite{hariat2026ss3d}  & \checkmark & $<20$sec & 0.090 & 0.043 & 0.601 & 0.092 & 0.026 & 1.064 \\
& \textbf{Ours: SS3D + VLRC} & \checkmark &  $<20$sec & \textbf{0.088} & \textbf{0.041} & 0.587 & \textbf{0.090} & \textbf{0.021} & \textbf{1.038} \\
\bottomrule
\end{tabular}
\caption{Pose estimation comparison to AnyCam\cite{wimbauer2025anycam} in zero shot. Absolute trajectory error (ATE) and relative pose error for translation (RPEtrans)
and rotation (RPErot) on the Sintel and TUM-RGBD datasets. }
\label{table:results_pose}
\end{table*}

\begin{table*}[!t]
    \centering
    \small
    \renewcommand{\arraystretch}{1.15}
    \begin{tabular}{lcc}
        \toprule
        \textbf{Method} & \textbf{AFE (px)$\downarrow$} & \textbf{RFE (\%)$\downarrow$} \\
        \midrule
        UniDepth & 447.4 & 35.7 \\
        Dust3r   & 434.0 & 36.4 \\
        AnyCam   & 252.2 & 18.1 \\
        SS3D & 256.6 & 16.7\\
        \textbf{Ours: SS3D + VLRC} & \textbf{255.5} & \textbf{16.5} \\
        \bottomrule
    \end{tabular}
    \caption{\textbf{Intrinsic parameter estimation on Sintel.} 
    Mean absolute focal error (AFE) and mean relative focal error (RFE).}
    \label{table:results_intrinsic}
\end{table*}

\paragraph{Intrinsic Estimation.}
Table~\ref{table:results_intrinsic} reports intrinsic estimation results. SS3D+VLRC predicts camera intrinsics directly from raw monocular videos, without external calibration cues or privileged information. Despite this fully self-supervised setting, our method matches AnyCam in absolute focal error (AFE) and achieves a substantially lower relative focal error (RFE). It also brings clear improvements over the original SS3D baseline.\\
Overall, the consistent gains across depth, pose, and intrinsics demonstrate that vision-language reprojection improves the core geometric quantities required for 3D reconstruction. Table~\ref{table:results_dino_clip} further shows that CLIP features outperform DINO features in our reprojection framework, suggesting that the language-aligned semantic information encoded by CLIP provides a stronger supervision signal than purely visual features.\\

We also provide qualitative results in Fig.~\ref{fig:qualitative_kitti}, showing predicted depth maps and the corresponding 3D reconstructions on several sequences of KITTI Odometry dataset. For clearer visualization, we remove sky regions using an off-the-shelf sky segmentation network before rendering the point clouds.

\section{Additional Implementation Details}
\label{sec:appendix:implementation_details}
We provide additional details for retraining SS3D+VLRC on YouTube8M.
We follow the same training protocol as the authors of SS3D\cite{hariat2026ss3d}. See the paper for more details. Here is an overview.\\
\textbf{Architecture}: we use the VGGT~\cite{wang2025vggt} architecture for our pipeline, keeping only the depth, pose and intrinsic heads.\\
\noindent
\textbf{Preprocessing.} Shot detection, frame-rate normalization, and frame filtering are performed with PyAV.\\
\textbf{Validity masking.} We use CoopNet~\cite{hariat2023rebalancing} to identify unreliable pixels, including occlusions and moving objects.\\
\noindent
\textbf{Training of student}: 
To construct each training batch, we first randomly select three clusters and then sample a total of 24 images from them. Instead of applying standard backpropagation, we follow the method proposed in \cite{sener2018multi} to compute Pareto-optimal gradients. During training, images are resized such that their shortest side is 518 pixels, after which a \(518 \times 518\) crop is extracted.\\
\noindent
\textbf{Hyperparameters}:
We used \(\lambda_{\text{distill}}\) = 0.2.

\paragraph{Web-video qualitative results.}
For the qualitative web-video results in the main paper, we use SS3D+VLRC retrained on YouTube8M~\cite{abu2016youtube} following the self-supervised SS3D training protocol, with VLRC added as an auxiliary feature-space reprojection loss. No 3D annotations, camera poses, or depth supervision are used during this retraining.

\end{document}